%% file: main.tex
\documentclass[10pt,conference]{IEEEtran}
\usepackage{cite}
\usepackage{amsmath,amssymb,amsfonts}
\usepackage{algorithmic}
\usepackage{graphicx}
\usepackage{textcomp}
\usepackage[hyphens]{url}
\usepackage{fancyhdr}
\usepackage{hyperref}
\usepackage{xspace}

\usepackage{comment}
\usepackage{multirow}
\usepackage[table,xcdraw]{xcolor}
\usepackage{colortbl}
\usepackage{soul}

\newcommand{\acceleratorname}{\textit{EpochCore}\xspace}
\newcommand{\dataflowname}{\textit{ProDF}\xspace}
\newcommand{\newPE}{\textit{LIMA-PE}\xspace}
\newcommand{\newPEs}{\textit{LIMA-PEs}\xspace}

\newcommand{\ignore}[1]{}

\newcommand{\revCommon}{\textcolor{black}}
\newcommand{\revA}{\textcolor{black}}
\newcommand{\revB}{\textcolor{black}}
\newcommand{\revC}{\textcolor{black}}
\newcommand{\revD}{\textcolor{black}}
\newcommand{\revE}{\textcolor{black}}
\newcommand{\revF}{\textcolor{black}}

\newcommand{\microAdd}{\textcolor{black}}
\newcommand{\labeltext}[3][]{%
    \@bsphack%
    \csname phantomsection\endcsname
    \def\tst{#1}%
    \def\labelmarkup{\emph}
    \def\refmarkup{}%
    \ifx\tst\empty\def\@currentlabel{\refmarkup{#2}}{\label{#3}}%
    \else\def\@currentlabel{\refmarkup{#1}}{\label{#3}}\fi%
    \@esphack%
    \labelmarkup{#2}
}

\title{Systolic Array-Based Accelerator For Structured State-Space Models }

\author{\IEEEauthorblockN{Shiva Raja\(^\dag\), Cansu Demirkiran\(^\dag\), Aakash Sarkar\(^{\ddag}\), Milo\v s Popovi\' c\(^{\dag}\), Ajay Joshi\(^\dag\)}  \IEEEauthorblockA{\textit{\(^{\dag}\)ECE Dept, \(^{\ddag}\) Psychological and Brain Sciences, Boston University} }}

\begin{document}

\maketitle
\pagestyle{plain}

\input{abstract}

\input{introduction_2}

\input{background}

\input{sithcore_architecture}

\input{evaluation}
\input{related_work}

\input{conclusion}

\section*{ACKNOWLEDGMENTS}
We want to acknowledge Prof Marc Howard for his valuable contributions to the initial ideas that helped shape the foundation of this paper. His insights were instrumental in guiding the early stages of this work.


\bibliographystyle{IEEEtranS}
\bibliography{main}


\end{document}

%% file: abstract.tex
\begin{abstract}
Sequence modeling is crucial for AI to understand temporal data and detect complex time-dependent patterns. 
While recurrent neural networks (RNNs), convolutional neural networks (CNNs), and Transformers have advanced in capturing long-range dependencies, they struggle with achieving high accuracy with very long sequences due to limited memory retention (fixed context window). 
State-Space Models (SSMs) leverage exponentially decaying memory enabling lengthy context window and so they process very long data sequences more efficiently than recurrent and Transformer-based models. 
Unlike traditional neural models like CNNs and RNNs, SSM-based models require solving differential equations through continuous integration, making training and inference both compute- and memory-intensive on conventional CPUs and GPUs. 

In this paper we introduce a specialized hardware accelerator, ~\emph{\acceleratorname}\footnote{Epoch does not refer to training iterations, instead it is an acronym.}, for accelerating SSMs.
\acceleratorname is based on systolic arrays (SAs) and is designed to enhance the energy efficiency and throughput of inference of SSM-based models for long-range sequence tasks. 
Within the SA, we propose a versatile processing element (PE) called \emph{LIMA-PE} to perform traditional and specialized MAC operations to support traditional DNNs and SSMs. 
To complement the \acceleratorname microarchitecture, we propose a novel dataflow, \emph{ProDF}, which enables highly efficient execution of SSM-based models. 
By leveraging the \textit{LIMA-PE }microarchitecture and \textit{ProDF}, \emph{\acceleratorname} achieves on average \(\sim 2000\times\) improvement in performance on LRA datasets compared to a GPU and
\(~250\times\) gains in performance and \(~45\times\) improvement in energy efficiency, 
over traditional SA-based accelerators (TPU). 
\vspace{-0.1in}
\end{abstract}

%% file: introduction_2.tex
\section{Introduction}\label{Intro}
The ability to detect and model patterns in extremely long input sequences is critical for achieving high accuracy in modern machine learning tasks such as large language modeling \cite{BidirectionLSTM}, video processing \cite{Seq2SeqVideoToText}, and speech recognition \cite{DualPathRNN}. 
However, the effectiveness of conventional architectures—including RNNs (e.g., LSTM \cite{LSTM_Paper}, GRU \cite{GRU}), and attention-based Transformers \cite{Attention}—is fundamentally constrained by scalability bottlenecks. 
These models typically struggle to handle input lengths beyond 16K tokens \cite{SSM2}, leading to degraded performance and computational inefficiencies on long-context tasks.

\begin{table}[]
\caption{Long Range Area (LRA) Dataset SOTA Performance. PP = Perplexity Percentage}
\vspace{-0.1in}
\label{tab:lra-datasets}
\begin{tabular}{llll}
\hline
\rowcolor[HTML]{C0C0C0} 
\textbf{\begin{tabular}[c]{@{}l@{}}LRA \\ Task\end{tabular}} & \textbf{\begin{tabular}[c]{@{}l@{}}Seq.\\ Length\end{tabular}} & \textbf{\begin{tabular}[c]{@{}l@{}}Other\\ Models'\\ Accuracy \%\end{tabular}} & \textbf{\begin{tabular}[c]{@{}l@{}}SSM based\\ Models'\\ Accuracy \%\end{tabular}} \\ \hline
Long ListOps \cite{LIQ_SSM1} & 2048 &\begin{tabular}[c]{@{}l@{}}49.53\\(H-Trans.)\end{tabular}& 62.75 (Liq-S4) \\ \hline
\begin{tabular}[c]{@{}l@{}}Byte-level Text\\ Classification \cite{SSM2}\end{tabular} & 2048 &\begin{tabular}[c]{@{}l@{}}65.90\\(L-Trans.)\end{tabular}& 86.82 (S4) \\ \hline
\begin{tabular}[c]{@{}l@{}}Byte-level Doc\\ Retrieval \cite{SSM2}\end{tabular}& 4000 &\begin{tabular}[c]{@{}l@{}}79.56\\(N-former)\end{tabular}& 90.90 (S4) \\ \hline
\begin{tabular}[c]{@{}l@{}}Image \\ Classification \cite{SSM2}\end{tabular} & 1024 & \begin{tabular}[c]{@{}l@{}} 47.38\\ (Luna-256)\end{tabular}& 88.65 (S4) \\ \hline
Pathfinder \cite{LIQ_SSM1} & 1024 & 91.70 (CDIL) & 94.8 (Liq-S4) \\ \hline
Pathfinder-X \cite{LIQ_SSM1} & 16384 & - & 96.66 (Liq-S4) \\ \hline
IMDB \cite{LIQ_SSM1} & 2048 & 86.78 (CDIL) & 89.02 (Liq-S4) \\ \hline
AAN \cite{LIQ_SSM1} & 4000 & 85.36 (CDIL) & 91.20 (Liq-S4) \\ \hline
sCIFAR \cite{LIQ_SSM1} & 3072 & \begin{tabular}[c]{@{}l@{}} 80.82 \\(FlexConv)\end{tabular} & 92.02 (Liq-S4) \\ \hline
PG-19 \cite{SSM4} & 65K & - & PP 12.47 (GSS) \\ \hline
Arxiv \cite{SSM4} & 65K & - & PP 2.75 (GSS) \\ \hline
Github \cite{SSM4} & 65K & - & PP 2.12  (GSS) \\ \hline
\end{tabular}
\vspace{-0.25in}
\end{table}

To address this limitation, recent research has explored alternative architectures based on State Space Models (SSMs)~\cite{ssm_survey_2024}, which offer a promising framework for scalable sequence modeling. 
The Linear State Space Layer (LSSL) \cite{SSM3} demonstrated early improvements over both RNNs and Transformers on simpler sequence tasks. Further enhancements to SSMs introduced structured state transition matrices—e.g., in S4 \cite{SSM2}, S5 \cite{SSM_S5}, DSS \cite{DSS}, and H3 \cite{H3}—to boost both accuracy and efficiency. 
More recent models, such as Liquid-S4 \cite{LIQ_SSM1, LIQ_SSM2}, GSS \cite{SSM4}, and Mamba \cite{gu2023mamba, li2025mamba}, integrate time-varying parameters and gating mechanisms, achieving state-of-the-art results on long-range benchmarks such as the Long Range Arena (LRA)~\cite{LRA} as shown in Table~\ref{tab:lra-datasets}.
These benchmarks encompass a wide range of applications, including 500+ token language tasks, 1000+ time steps for time-series data, and hundreds of video frames. For example, Gu et al.~\cite{SSM2} report that S4 achieves a 5.19\(\times\) speedup and 0.091\(\times\) memory usage compared to Transformers on 4K-length inputs, while also delivering over 60\(\times\) throughput gains in token processing.  
\revCommon{SSM is not suited for non-sequential data such as grid-based data, graph-structure, recommendation systems, structured reasoning etc.~\cite {IllusionSSM}. However, we can transform non-sequential data into sequential data through an offline preprocessing step \cite{ssm_survey_2024}. and then use SSM for processing it.}

\begin{figure}[!ht]
    \centering
\includegraphics[width=0.48\textwidth]{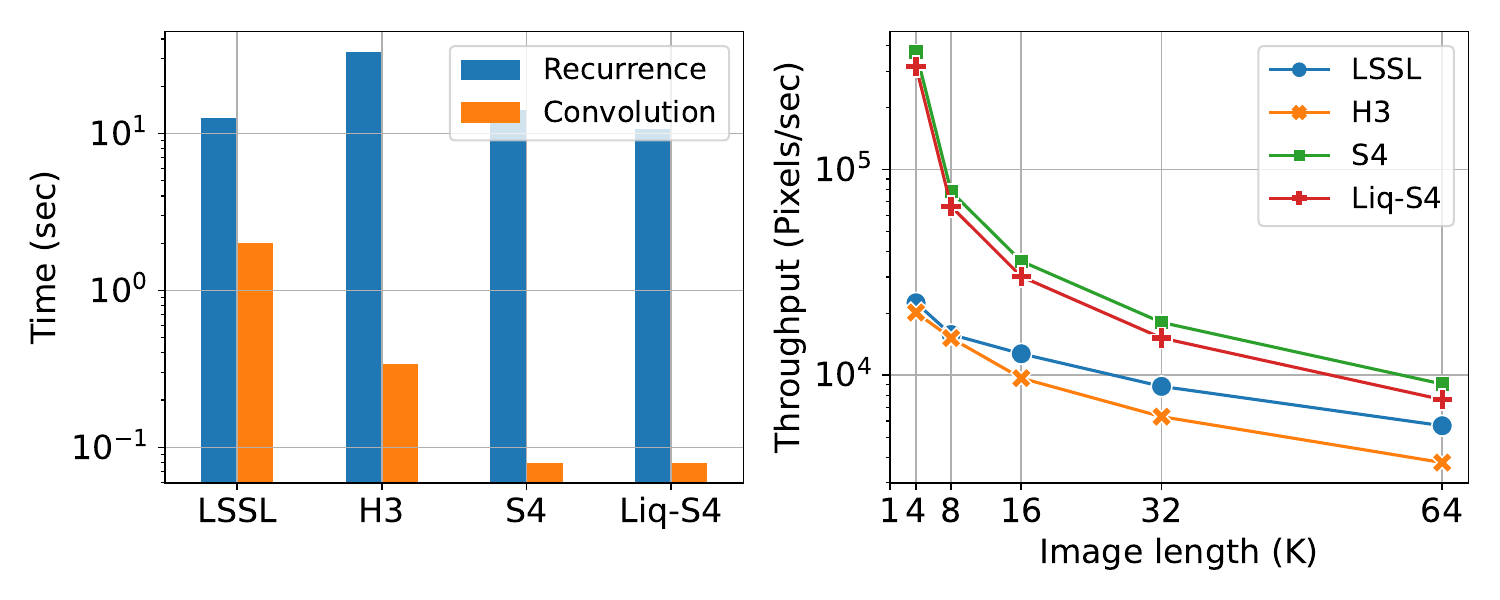}
    \vspace{-0.15in}
     \caption{(Left) GPU (Nvidia  A100) latency of SSM models for recurrence vs convolutional methods when processing images with 4K sequence length and batch size of 64. (Right) For convolution, the throughput per pixel decreases over sequence lengths.}
    \label{fig:SsmEasyGpus}
    \vspace{-0.2in}
\end{figure}

Despite these algorithmic advances, hardware platforms have not kept pace. General-purpose GPUs and TPUs rely heavily on vectorized GEMM operations, which are ill-suited for the sparse, recurrent, and long-kernel convolution computations that define SSM-based models. 
GPU-based FFT accelerations, such as cuFFT and tcFFT~\cite{tcFFT}, improve throughput but still fall short in energy efficiency and memory locality. 
Prior custom architectures like Boriakoff’s 1D systolic array ~\cite{FFT_SA} offer specialized acceleration for FFT-based convolutions but cannot generalize to the broader computational patterns in SSMs or traditional DNN layers. 
Additionally, as convolutional SSM implementations scale to longer sequences, they become prohibitively memory-intensive and exhibit sharp drops in throughput (see Figure \ref{fig:SsmEasyGpus}). 
SSM-based models often rely on intermediate coefficients represented as complex numbers, introducing extra computational overhead and increasing hardware complexity on custom accelerator architectures.

To meet these emerging computational demands, we propose \emph{\acceleratorname} (ExPOnentially-Compressed History Core), a digital hardware accelerator designed to efficiently execute both structured SSM models (e.g., S4, Liquid-S4) and traditional dense neural networks (e.g., CNNs, RNNs, Transformers). 
\acceleratorname is built using an array of a novel Processing Element (PE) called \newPE, which supports multiple MAC operation modes. 
These include recurrent integration with fixed (S4) and time-varying (Liquid-S4) coefficients, as well as standard GEMM dataflows for traditional DNN layers. 
\newPE also has a unified design to process either real or complex numbers for the MAC operations. 
To reduce dynamic power, each \newPE employs dual gated clocks to selectively minimize switching activity.

We also propose a novel programmable dataflow  (\emph{\dataflowname}) that enables efficient pipelined execution of both sparse and dense matrix operations on \acceleratorname. 
In addition to standard north-south and west-east flows, \dataflowname incorporates a novel northeast-southwest dataflow to accelerate elementwise recurrent computations within SSM layers.

Our work makes the following key contributions:
\begin{enumerate}
\item \emph{\acceleratorname} Architecture: We introduce \acceleratorname, the first hardware accelerator that natively executes multiple Structured SSMs (e.g., S4, Liquid-S4), while retaining support for traditional DNNs. 
Effectively, we have a unified inference system for both long-sequence and conventional models.

\item \emph{\newPE} Design: We design a versatile PE capable of operating in four MAC modes: Fixed Recurrent Integration (FRI-MAC), Time-Varying Recurrent Integration (TRI-MAC), Banded Weight Stationary (BWS-MAC), and Traditional Output Stationary (TOS-MAC). \newPE also has an efficient and unified MAC circuitry to support real and complex valued data types.  Dual gated clocks enhance energy efficiency by decoupling load and compute phases.

\item Programmable Dataflow (\emph{\dataflowname}): We propose \dataflowname, a novel dataflow that exploits pipelining across unconventional directions to support both sparse recurrent updates and dense GEMM computations.
\end{enumerate}

We evaluated \emph{\acceleratorname} on multiple LRA datasets, comparing it against three SoTA general-purpose accelerators: (1) a 1D systolic array optimized for long-kernel convolutions~\cite{FFT_SA}, (2) a traditional SA with sparsity support~\cite{STPU}, and (3) GPUs. 
Compared to the 1D FFT-accelerated SA, run on S4 and Liquid-S4 models, \textit{\acceleratorname} delivers on an average 25\(\times\) speedup, 10\(\times\) energy savings, and 30\(\times\) lower memory bandwidth.
For S4 and Liquid-S4 models on LRA datasets, \acceleratorname achieves up to 250\(\times\) performance improvement and 45\(\times\) energy reduction compared to traditional PE-based SAs. 
For inference latency on S4 layer and Liquid-S4 layers, \acceleratorname outperforms GPUs by a factor of 2000\(\times\). 

We also compared \acceleratorname's performance improvements over GPUs against three recently proposed SSM accelerators: (1) For H3~\cite{H3} model inference, \acceleratorname achieves \(3860\times\) improvement, while the VGA~\cite{VGA} accelerator shows a \(4\times\) improvement over a GPU. (2) For Mamba model inference, \acceleratorname provides a \(4.75\times\) improvement, while on Marca~\cite{li2024marca} achieves an \(11.66\times\) and FastMamba~\cite{FastMamba} achieves a \(6.06\times\) speedup over a GPU.

%% file: background.tex

\section{Background}
In this section, we provide the background for 
the recently proposed Structured State-Space Sequential Models (S4). We also discuss the extension of S4 to include input-dependent time-varying coefficients, known as Liquid-S4. 
These models rely on mapping inputs to internal state parameters, which are governed by a first-order ordinary differential equation (ODE).
\vspace{-0.1in}

\subsection{State-Space Models (SSM)}\label{BG_SSM}


SSMs are designed to capture long-range dependencies by approximating the input sequence using coefficients mapped to an orthogonal polynomial (OP) basis. This approximation is implemented in a framework called High-order Polynomial Projection Operators (HiPPO)\cite{HIPPO1}. 

In an SSM, a 1-D input sequence $u(t) {\in} \mathbb{R}$ is mapped to an internal state $\mathbf{x}(t){\in} \mathbb{R}^{N\times 1}$ at every position in the sequence. This internal state is updated according to a linear first-order ODE:
\vspace{-0.05in}
\begin{equation}\label{SSM_Eqn1}
\frac{d}{dt} \mathbf{x}(t)=\mathbf{A}\cdot \mathbf{x}(t)+\mathbf{B}\cdot u(t)
\end{equation}
where $\mathbf{A}\in \mathbb{R}^{N\times N}$ and $\mathbf{B}\in \mathbb{R}^{N\times 1}$ are initialized based on the chosen OP basis and are updated during training. The output sequence $y(t) \in \mathbb{R}$ is obtained via a linear transformation of the internal state and input:
\vspace{-0.05in}
\begin{equation}\label{SSM_Eqn2}
y(t)=\mathbf{C}\cdot \mathbf{x}(t)+\mathbf{D}\cdot u(t)
\end{equation}
where $\mathbf{C}\in \mathbb{R}^{1\times N}$ and $\mathbf{D}\in \mathbb{R}$ are fully trainable coefficients.

Recent studies \cite{SSM1, SSM2, SSM3, LIQ_SSM1} highlight the effectiveness of using the Scaled-Legendre (HiPPO-LegS) OP basis for improved accuracy. However, diagonalizing the coefficient matrix $\mathbf{A}$ for HiPPO-LegS can be computationally challenging.
To address this, Gupta et al. \cite{DSS} proposed the simpler Diagonal State Space (DSS) model, which computes the complex eigenvalues of the non-scaled HiPPO matrix. 
The eigenvalue-based diagonalization of the HiPPO matrix simplifies computation, making it an efficient and attractive approach as shown by Gu et al in S4 ~\cite{SSM2}. 

The recurrent update for the internal state in this diagonalized formulation is given as:
\vspace{-0.05in}
\begin{equation}\label{ODE_dss_final}
\mathbf{x}(t+\triangle t)=\mathbf{\overline{A}}\odot\mathbf{x}(t)+\mathbf{\overline{B}}\cdot u(t)
\end{equation}
where $\Lambda \in \mathbb{C}^{N\times N}$ is a diagonal matrix with eigenvalues $\lambda_0,...,\lambda_N$ , $\forall i, \lambda_i \neq0$, the coefficient matrices given by $\mathbf{\overline{A}}=diag[e^{-\mathbf{\Lambda}\triangle t}]\in \mathbb{C}^{N\times 1}$, $\mathbf{\overline{B}}=diag[\mathbf{\Lambda}^{-1}(I-e^{-\mathbf{\Lambda}\triangle t})\mathbf{B}]\in \mathbb{C}^{N\times 1}$ and $\odot$ denotes element-wise multiplication.
This formulation maintains a structure similar to the earlier recurrence 
while leveraging eigenvalues for computational simplicity, however introducing complex valued coefficients. 

\textbf{Liquid-S4:}
\label{BG_LIQ_SSM}
By introducing a dynamic, time-varying (or \emph{liquid}) time-constant coefficient into the internal state dynamical equation \eqref{SSM_Eqn1}, Hasani et al, demonstrated significant improvements in time-series prediction accuracy~\cite{LIQ_SSM1, LIQ_SSM2}. 
The updated dynamic equation for the internal state is expressed as:
\vspace{-0.05in}
\begin{equation}\label{LiqSSM_Eqn1}
\frac{d}{dt} \mathbf{x}(t)=[\mathbf{A}+\mathbf{B}\cdot u(t)]\cdot \mathbf{x}(t)+\mathbf{B}\cdot u(t)
\end{equation}
where $\mathbf{A}$ and $\mathbf{B}$ are matrices that define the state-space dynamics, and \(u(t)\) is the input signal. The inclusion of a dynamic coefficient allows the system to adapt its internal state evolution based on the input. Using the bilinear transform for discretization, the solution to Equation \eqref{LiqSSM_Eqn1} becomes:
\vspace{-0.05in}
\begin{equation}\label{LiqSSM_Eqn2}
\mathbf{x}(t+\triangle t)=[\mathbf{\overline{A}}+\mathbf{\overline{B}}\cdot u(t)]\odot\mathbf{x}(t)+\mathbf{\overline{B}}\cdot u(t)
\end{equation}
where the matrices $\mathbf{\overline{A}}$ and $\mathbf{\overline{B}}$ are computed in the same manner as in S4, and $\odot$ represents element-wise multiplication. This formulation enables the model to adaptively adjust its dynamics, enhancing its ability to process time-varying signals effectively.

\textbf{Solving SSMs}:
Two widely used approaches for solving the combined Equation \eqref{ODE_dss_final} and \eqref{SSM_Eqn2} are  the convolution and recurrent methods \cite{SSM3}.

\subsubsection{Convolution Method for Solving SSMs}
The convolution method involves computing the 1-D output sequence \(y(t)\) by applying a non-circular convolution of the discretized Krylov function, \(\kappa_L(C,A,B)=(C A^i B)_{i\in L} \), over the entire input sequence. This technique is particularly suitable for batch processing scenarios, such as training, where the entire input sequence is available upfront. 
However, this method has notable drawbacks. (a)  \emph{High Memory Usage}: It requires approximately 144\(\times\) the size of the input sequence for both inference and training. (b) \emph{Kernel Generation Overhead}: The computational cost of generating the full kernel increases with both the sequence length and the size of the state map. 

\textbf{Hardware Accelerators}: A specific decomposition of the Cooley-Tukey matrix enables an efficient implementation using linearly connected systolic arrays, as demonstrated by Boriakoff in~\cite{FFT_SA}. 
Despite its efficiency and suitability for specific applications, this architecture has limitations in scaling to different sequence lengths and supporting GEMM computations.
\begin{figure}[t]
    \centering
    \includegraphics[width=0.45\textwidth]{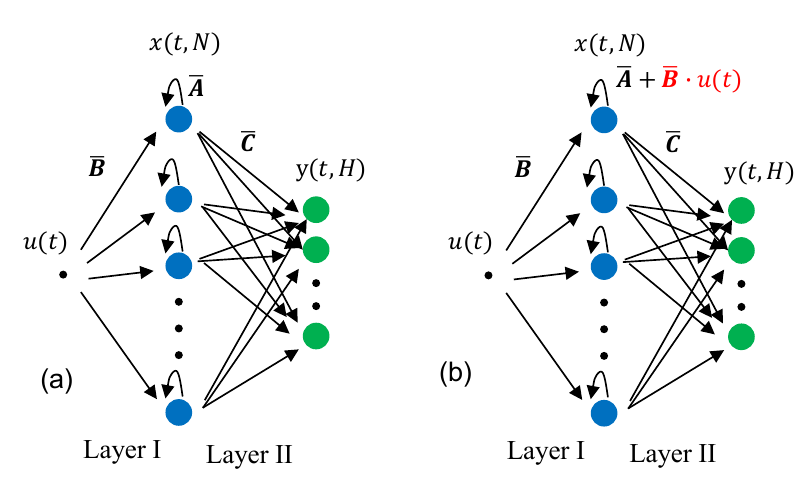}
    \vspace{-0.15in}
    \caption{(a) S4- \cite{DSS}, (b) Liq-S4-based \cite{LIQ_SSM2} model with input sequences \(u(t)\) and output sequences \(y(t,H)\). Layer I determines the internal state space map recurrently. Layer II maps the internal state to the output sequence via a dense linear layer.}
    \label{fig:Hippo_NN_Layer}
    \vspace{-0.2in}
\end{figure}

\subsubsection{Recurrent Method for Solving S4}
In the recurrent model, the internal state vector, \(\mathbf{x}(t)\), is updated continuously at fixed time steps, making it ideal for real-time processing of potentially unbounded sequences, such as during inference. Figure \ref{fig:Hippo_NN_Layer} illustrates the equivalent neural network representation for S4-based and Liquid-S4-based models.

In Figure \ref{fig:Hippo_NN_Layer}, Layer I is governed by Equation \eqref{ODE_dss_final}. The 1-D input \(u(t)\) is scaled by the vector \(\mathbf{\overline{B}}\) to update a set of recurrent operations, each initialized with fixed coefficients defined by the vector \(\mathbf{\overline{A}}\). In Liquid-S4 models (governed by Equation \eqref{LiqSSM_Eqn2}), an additional input-dependent, time-varying coefficient \(\mathbf{\overline{B}}\cdot u(t)\) is included. Layer II, shown in Figure \ref{fig:Hippo_NN_Layer}, follows Equation \eqref{SSM_Eqn2} and is implemented as a standard matrix multiplication operation, commonly referred to as the linear layer.  

\textbf{Hardware Accelerators}: 1 Layer I computations to GEMM operations on 2-D SAs, such as TPUs, is not straightforward. 
While the linear scaling \(\mathbf{\overline{B}}\cdot u(t)\) can be implemented via diagonal matrix-vector multiplication, there are no native MAC operations for recurrent updates. 
To address this, we introduce a new cardinal MAC operation for recurrent updates, defined as: 
\vspace{-0.1in}
\begin{equation}\label{ODE_mac}
x\Leftarrow a\cdot x+b
\end{equation}
For Liquid-S4 models with time-varying coefficients, Equation \eqref{LiqSSM_Eqn2} can also be mapped to this new cardinal MAC operation with an additional summation step:
\vspace{-0.1in}
\begin{equation}\label{ODE_mac_liq}
x\Leftarrow (a+b)\cdot x+b
\end{equation}
Moreover, the coefficients \(\mathbf{\overline{A}}\), \(\mathbf{\overline{B}}\) and the internal state vector \(\mathbf{x}(t)\) can be complex valued data.
In section \ref{EpochArchSection}, we detail the modifications required in traditional processing elements (PEs) to support these new cardinal MAC operations, as well as support either real or complex valued data.

Layer II computations can be directly mapped to GEMM operations of a TPU. 
The overall pipelining and throughput of the SSM model evaluation depend on the dataflow approach. 
In this paper, we recommend a variant of weight-stationary dataflow, which is particularly suitable for longer input and output sequences that move through the SA efficiently.

\begin{figure}[b]
    \centering
    \vspace{-0.2in}
    \includegraphics[width=0.4\textwidth]{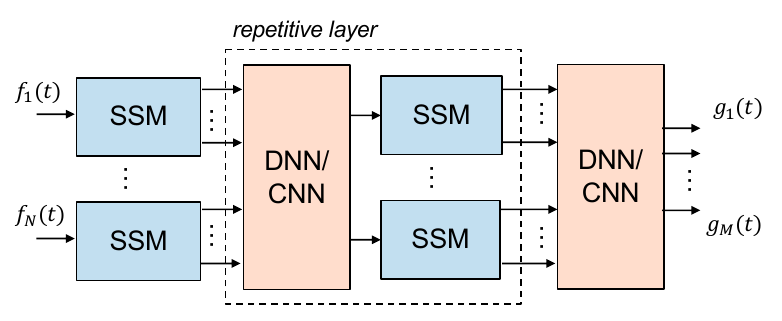}
    \vspace{-0.15in}
    \caption{SSM-based models are multi-layered with interleaved SSM and DNN/CNN layers that process input sequences \(f(t)\) and generate output sequences or labels \(g(t)\). }
    \label{fig:SithSsmMultiLayer}
\end{figure}

\textbf{Deep SSM:}
The SSM based models, illustrated in Figure \ref{fig:Hippo_NN_Layer}, follows a defined sequence of operations outlined in Table \ref{tab:epoch_layer_steps}. 
Each SSM model can be configured to produce multi-dimensional output sequences, called heads (\(H\)), with each head trained to extract distinct features from the 1-D input sequence. These SSM model layers can be interleaved with DNN or CNN layers, as shown in Figure \ref{fig:SithSsmMultiLayer} to enhance their functionality. 

Gu et al. \cite{SSM2} and Hasani et al. \cite{LIQ_SSM1}, have demonstrated SoTA accuracy for various LRA datasets by interleaving up to six SSM model layers with DNN layers. 
This multi-layer configuration, shown in Figure \ref{fig:SithSsmMultiLayer}, enhances the model's ability to process long-range dependencies. 
The \acceleratorname accelerator, discussed in section \ref{EpochArchSection}, is designed to efficiently execute both SSM and non-SSM model layers such as DNN and CNNs, providing a versatile platform for diverse deep learning workloads.

\begin{table*}[h!]
\caption{Operations in the Structured SSM Layer}
\centering
\label{tab:epoch_layer_steps}
\begin{tabular}{llll}
\hline
\rowcolor[HTML]{C0C0C0} 
\textbf{\begin{tabular}[c]{@{}l@{}} Steps\end{tabular}} & \textbf{\begin{tabular}[c]{@{}l@{}}Cycles\end{tabular}} & \textbf{\begin{tabular}[c]{@{}l@{}}S4 Layer\end{tabular}} & \textbf{\begin{tabular}[c]{@{}l@{}}Liq-S4 Layer\end{tabular}} \\ \hline
Scale Input & 1 & \begin{tabular}[c]{@{}l@{}}\(\mathbf{B}.u(t)\)\end{tabular} & \begin{tabular}[c]{@{}l@{}}\(\mathbf{B}.u(t)\)\end{tabular} \\ \hline
\begin{tabular}[c]{@{}l@{}} State  Equation\end{tabular} &  & \begin{tabular}[c]{@{}l@{}}\(\frac{d}{dt}\mathbf{X}(t) \)\(= \mathbf{A}\odot \mathbf{X}(t)+\mathbf{B}\cdot u(t)\)\end{tabular}
& \begin{tabular}[c]{@{}l@{}}\(\frac{d}{dt}\mathbf{X}(t) \)\(= [\mathbf{A}+\mathbf{B}\cdot u(t)]\odot \mathbf{X}(t)+\mathbf{B}\cdot u(t)\)
\end{tabular} \\
\begin{tabular}[c]{@{}l@{}}Recurrent Integration\end{tabular} & 1 & \begin{tabular}[c]{@{}l@{}}\(\mathbf{X}(t_+)= \mathbf{\overline{A}}\odot \mathbf{X}(t_-)+\mathbf{\overline{B}}\cdot u(t)\)\end{tabular}
 & \begin{tabular}[c]{@{}l@{}}\(\mathbf{X}(t_+)= [\mathbf{\overline{A}}+\mathbf{\overline{B}}\cdot u(t)]\odot \mathbf{X}(t_-)+\mathbf{\overline{B}}\cdot u(t)\)
\end{tabular} \\ 
\hline
\begin{tabular}[c]{@{}l@{}}Linear Layer\end{tabular} & N & \begin{tabular}[c]{@{}l@{}}\(\mathbf{y}(t)=\mathbf{C}\times \mathbf{X}(t)\)\end{tabular} 
& \begin{tabular}[c]{@{}l@{}}\(\mathbf{y}(t)=\mathbf{C}\times \mathbf{X}(t)\)\end{tabular}
\\ \hline
\end{tabular}
\vspace{-0.15in}
\end{table*}

%% file: sithcore_architecture.tex
\section{\acceleratorname Architecture} \label{EpochArchSection}

\begin{figure}[t]
    \centering
    \includegraphics[width=0.47\textwidth]{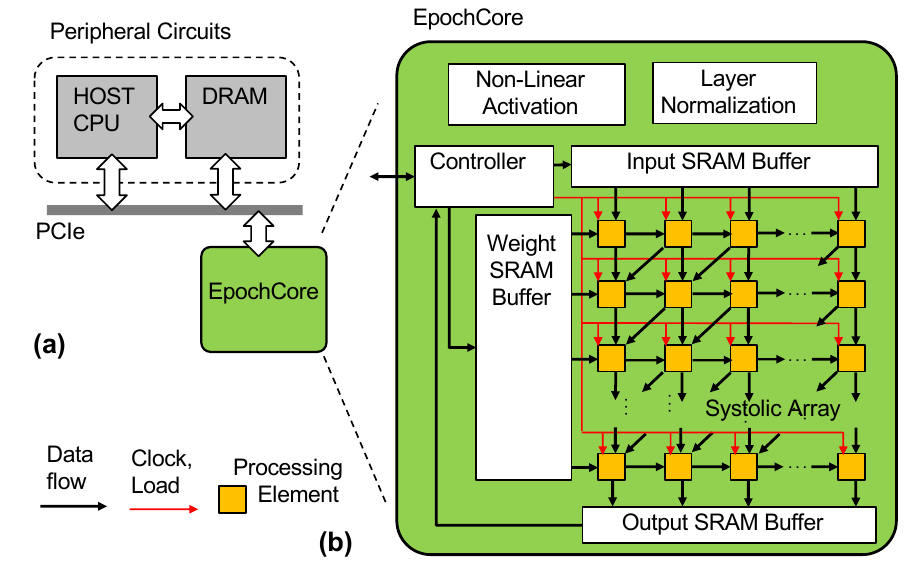}
    \vspace{-0.15in}
    \caption{(a) Architecture of the full System that uses \acceleratorname and (b) \acceleratorname micro-architecture. The \acceleratorname supports SSM and GEMM operations.}
    \label{fig:EpochCoreMicroArch}
    \vspace{-0.25in}
\end{figure}

This section outlines the microarchitecture of the  \acceleratorname accelerator.
Similar to the TPU, the \acceleratorname is implemented on a standalone card interfacing with the CPU and DRAM through a PCI interface, as shown in Figure \ref{fig:EpochCoreMicroArch}a. 
The host CPU offloads instructions to the \acceleratorname and manages data transfer between the CPU and the accelerator. Internally, the Controller Unit manages the dataflow within the \acceleratorname. 
We use specialized on-chip hardware in \acceleratorname for non-linear and normalization operations.

\subsection{\acceleratorname Micro-architecture}
\label{EpochMicroArchSection}
The \acceleratorname microarchitecture, shown in Figure \ref{fig:EpochCoreMicroArch}(b), consists of a 2D SA of \newPEs (see Section~\ref{LimaMicroArchSection}), specialized for \microAdd{Structured-SSM}
operations present in the S4 and Liquid-S4 models, as well as MAC operations in standard GEMM computations. 
The SA interacts with the on-chip SRAM units for memory operations. 
Two on-chip SRAM units are used: one for storing and accessing input and output data, and another for storing weights and control signals.

LRA datasets for Structured-SSM (S4 and Liquid S4) computations involve input and output sequences of length ranging from thousands to millions, making input-stationary (IS) or output-stationary (OS) approaches highly inefficient due to the need for extensive tiling. 
Since the weight matrix is the smallest among the weight, input or output matrices, a weight-stationary (WS) dataflow offers the lowest power consumption and highest throughput, while minimizing tiling needs.
Hence, \acceleratorname adopts a modified WS dataflow for Structured-SSM computations. The modification is in programatically altering the flow of output through the processing elements while keeping the weights stationary.
Preloading weights for an entire tile of size \(n\times n\) requires significant on-chip SRAM bandwidth.
\acceleratorname's unified dataflow for evaluating Structured-SSM in a single SA computation allows weights to be stationary throughout the processing of a full batch of inputs, thereby amortizing the high on-chip SRAM bandwidth for loading weights.
During Structured-SSM operations, the weights remain stationary while the 1D input sequence is broadcast across \(n\) columns of the top row of the 2D SA, where \(n\) is the size of the internal state-map of the S4 model. 


The \newPEs in the SA are capable of performing MAC operations (for both S4s and traditional DNNs) and storing weights and the intermediate results from recurrent computations. 
The \newPEs are interconnected, enabling data movement within the SA in West-to-East or North-to-South directions. 
Additionally,  
\acceleratorname supports diagonal data movement in the Northeast-to-Southwest direction, facilitating banded matrix multiplication and non-staggered multiplications required for Layer II (see Figure \ref{fig:Hippo_NN_Layer}) computations in S4 models. 
The \newPE is designed with a specific bit-precision. It supports data representation in either real or complex fixed-point formats. For complex values, the representation is achieved by sharing the higher-order bits to store the real part, while using lower-order bits store the imaginary part. As a result, processing complex values effectively halves the bit-precision compared to real-valued data.

 
The controller unit decodes instructions from the host CPU to manage synchronized clocking, loading, and reset operations of the \newPEs. 
To enable the novel programmable dataflow, \dataflowname (discussed later in this section), the controller configures specific modes for each \newPE, facilitating the unique dataflow required to evaluate the S4 model for each tile. The three control input bits are multiplexed into the weight bus, with both the control bits and weights stored in the weight SRAM.
The \newPEs support both traditional MAC operations and specialized MAC operations needed for SSM recurrent computations. 
Each \newPE contains two internal buffers sets, implemented as registers: one for stationary operands, such as the weight matrices and control inputs that remain static during SA operations, and another for intermediate computed values, such as state-map vectors, which may move dynamically through the SA. 
\acceleratorname execution involves the following sequential operating phases:

\textbf{Reset Phase}:
This phase initializes all the \newPE by clearing all PE internal buffers, where we store the stationary and control inputs, and the intermediate outputs.

\textbf{Pre-Load Phase}: 
During this phase, each \newPE loads the control and weight data into the \newPE control and stationary buffers, respectively. 
These inputs, sourced from on-chip weight SRAM, prepare the PEs for subsequent operations.

\textbf{Compute Phase}: 
In this phase, data flows through the PEs to evaluate the S4, Liquid-S4 or the traditional DNN model. 
The process is pipelined to enhance throughput. 
Depending on what is suitable for the S4, Liquid-S4 or DNN model, one can use input stationary, weight stationary or output stationary dataflow in \acceleratorname.

\textbf{Readout Phase}:
The final output is stored in multiple rows of \newPE.
After the completion of the entire matrix computation, the outputs need to be sequentially read out onto the on-chip output SRAM during the readout phase. 


\vspace{-0.05in}
\subsection{\newPE Micro-architecture}\label{LimaMicroArchSection}
\begin{figure}[t]
    \centering
    \includegraphics[width=\columnwidth]{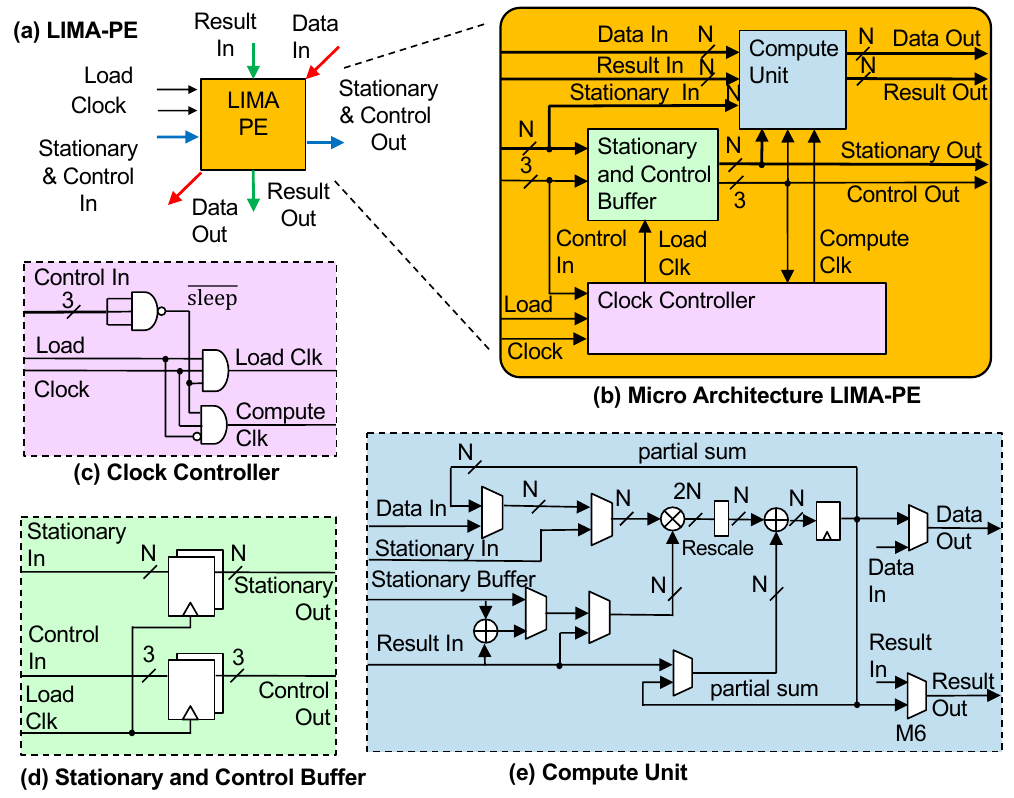}
     \vspace{-0.25in}
    \caption{\newPE Design: (a) Input flow directions. (b) Micro architecture details. (c) Gated clock circuitry for energy efficiency (d) Buffer for stationary and control inputs (e) Mode-specific MAC computation with a buffer for the partial result.}
    \label{fig:LimaMicroArch}
    \vspace{-0.3in}
\end{figure}

\begin{figure*}[ht]
    \centering
    \vspace{-0.15in}
    \includegraphics[width=0.9\textwidth]{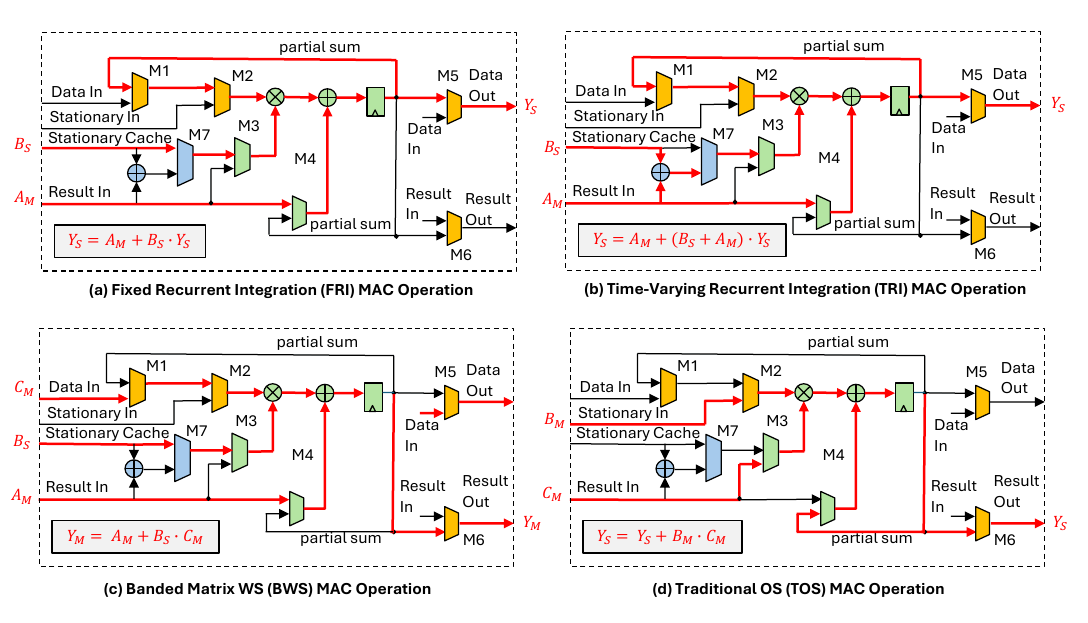}
    \vspace{-0.25in}
    \caption{Data flow under various MAC operating modes of LIMA-PE. The \emph{green} logic are part of traditional-PE design. LIMA-PE includes additional \emph{yellow} logic for recurrent integration and banded matrix dataflows, and \emph{blue} logic for time-varying integration support. The select inputs of the muxes, driven by the buffered control inputs, enable the operating mode of the LIMA-PE. (a) and (b) are novel additions to the LIMA-PE to support recurrent integrations for S4 and Liquid-S4 models. (c) is a modified WS MAC operation to enable diagonal dataflow in the SA for banded-matrix multiplication and (d) support regular GEMM for OS data flow. WS \& IS MAC operations, pass-through and sleep modes are not shown, but can be easily deduced.}
    \label{fig:LimaMacOperatingModes}
    \vspace{-0.15in}
\end{figure*}

The microarchitecture of the \newPE is illustrated in Figure \ref{fig:LimaMicroArch}.
Like traditional PEs, a \newPE is single-buffered and incorporates registers to hold stationary data in the stationary buffer and moving data in the output buffer of the compute unit. 
During the \emph{Pre-Load Phase}, stationary data is loaded into the registers.
No MAC operations are performed during the pre-load phase.
In the subsequent \emph{Compute Phase}, inputs and intermediate outputs are transferred across the SA as MAC operations process the data to compute final results. 

For energy efficiency, the \newPE generates two mutually exclusive internal clocks-\emph{Load Clock} for Pre-Load Phase and \emph{Compute Clock} for Compute Phase, in the clock controller circuit.
While introducing clock-gating adds extra circuitry, it is a widely used technique for reducing power consumption. 
\acceleratorname includes additional control bits, which must be programmatically managed to enable the mutually exclusive clocks.

The throughput of an S4 layer processing in EpochCore is primarily limited by input and output bandwidth of the on-chip SRAM (more details in Section~\ref{EVAL}). 
Double buffering, a common method for improving performance, may not improve performance in \acceleratorname because loading weight matrices from weight SRAM to the \newPEs typically accounts for only a small portion of total compute cycles. 
The sizes of the weight and input-output on-chip SRAM are 16MB, each determined by the S4 model size, and these are chosen to minimize off-chip DRAM access.

During the Pre-load phase of the \acceleratorname, a \newPE can operate in the following three modes:

\textbf{Accumulation Mode}:
In accumulation mode, the \newPE performs MAC operations similar to traditional PEs, guided by buffered input controls, as shown in Figure \ref{fig:LimaMicroArch}e. 
For the traditional GEMM operation in the WS, IS and OS dataflow, \newPE allows data to flow from \emph{Result In} to \emph{Result Out}, i.e., North-to-South. 
For the recurrent and banded-matrix MAC operations, dataflow is enabled additionally from \emph{Data In} to \emph{Data Out}, i.e., Northeast-to-Southwest.

\textbf{Pass-Through Mode}:
In this mode, the \newPE bypasses MAC operations, transferring data directly from inputs (\emph{Result In} and \emph{Data In}) to outputs (\emph{Result Out} and \emph{Data Out}) without accessing internal buffers.  In the absence of a Pass-Through mode, input from previous rows would need to be staggered across the columns adding more compute cycles.
The Pass-Through mode facilitates unaltered data transfer through the systolic array, optimizing the pipelining of matrix and banded-matrix multiplications.

\textbf{Sleep Mode}:
Sleep mode disables both MAC operations and data transfer. 
In this mode, the \emph{Load Clock} and \emph{Compute Clock} are turned off, significantly reducing energy consumption. 
\newPEs that are not utilized during the computation phase are set to Sleep mode.

\subsubsection{Traditional PE vs \newPE}
\label{sec:trad-PE-vs-LIMA-PE}
Traditional PEs are typically optimized to perform a specific MAC operation, which is tailored to support a designated dataflow. 
In OS dataflow, where the output \(Y_s\) remains stationary while the weight \(B_m\) and input \(C_m\) move, the Traditional-PE is designed to perform the following MAC operation:
\vspace{-0.1in}
\begin{equation}\label{OSMAC}
Y_s=B_m\cdot C_m+Y_s
\end{equation}
In contrast, for the WS and IS dataflows, the output \(Y_m\) moves while either the weight or input remains stationary. Traditional-PE supports the WS and IS dataflow through the MAC operation:
\begin{equation}\label{WSMAC}
Y_m=B_s\cdot C_m+Y_m
\end{equation}

Following are the novel additions to the LIMA-PE to support S4- and Liquid-S4-based models.

\textbf{Support multiple MAC types}: 
The \newPE is designed as a versatile architecture capable of supporting various MAC operations by dynamically controlling the dataflow through internal multiplexers. 
The configuration of these multiplexers (M1-M7), illustrated in Figure \ref{fig:LimaMacOperatingModes}, is determined by a 3-bit control input, which enables different types of MAC operations. 
\newPE supports the novel MAC operations (more information below) critical for recurrent integration in S4 and Liquid-S4 models.
Furthermore, it seamlessly performs the MAC operations defined in Equations \eqref{OSMAC} and \eqref{WSMAC}, facilitating WS, IS and OS dataflows for standard GEMM computations. 

\textbf{Fixed Recurrent Integration (FRI) MAC Operation}:
As detailed in section \ref{BG_SSM}, S4 models utilize Equation \eqref{ODE_dss_final} to describe the Linear-ODE solution for mapping an input sequence to an internal state vector through a discrete recurrent model. 
The corresponding MAC operation is expressed as:
\begin{equation}\label{LaplaceMAC}
Y_s=B_s\cdot Y_s+A_m
\end{equation}
where \(B_s \mapsto \mathbf{\overline{A}}\): the fixed recurrent coefficient, \(A_m \mapsto \mathbf{\overline{B}}\cdot u(t)\): the scaled input propagating through the SA in the North-to-South direction,  \(Y_s \mapsto \mathbf{x}(t)\): the current state vector (RHS), and \(Y_s \mapsto \mathbf{x}(t+\Delta t)\): the next state vector (LHS).
This operation effectively performs recurrent integration.
The MAC operation is implemented by incorporating multiplexers (M1, M2 and M5) and using buffered \emph{Control Out} signals to configure these multiplexers, as illustrated in Figure \ref{fig:LimaMacOperatingModes}a. 
Traditional PEs (such as TPUs) do not support such MAC operation.

\textbf{Time-Varying Recurrent Integration (TRI) MAC Operation}: For Liquid-S4 models, Equation \eqref{SSM_Eqn2} corresponds to the following MAC operation:
\begin{equation}\label{LiqMAC}
Y_s=(B_s+A_m)\cdot Y_s+A_m
\end{equation}
where \((B_s + A_m) \mapsto (\mathbf{\overline{A}}+\mathbf{\overline{B}}\cdot u(t))\): the time-varying coefficient for the recurrent integration operation.
This operation is implemented by incorporating an additional full-adder and the multiplexer M7, with the relevant circuitry highlighted in \emph{blue}, as shown in Figure \ref{fig:LimaMacOperatingModes}b. The inclusion of M7  introduces a novel mechanism to feed the scaled input into both the multiplier coefficient and the accumulator.
The TRI-MAC operation employs a dataflow approach similar to FRI-MAC, ensuring compatibility while supporting the time-dependent dynamics of Liquid-S4 models.

\textbf{Handling complex-valued data types:} 
As discussed in Section \ref{BG_SSM},  Equations \eqref{ODE_dss_final} and \eqref{LiqSSM_Eqn2} may involve complex-valued coefficients and state-map vectors. These complex values are represented using fixed-point format, where the real and imaginary components each occupy half the total bit width. Depending on a control bit, each \newPE can perform either a full-precision real-valued MAC operation or a half-precision complex-valued MAC operation. In the case of complex operations, all inputs and outputs of the \newPE are treated as complex values.

The compute unit within the \newPE, shown in Figure \ref{fig:LimaMicroArch}e, is designed to handle both real and complex fixed-point operations without requiring extensive specialization. Only the multiplication unit requires modification to support complex arithmetic. We designed the \newPE's multiplication unit to efficiently reuse sub-operand multiplications for both real and complex data, minimizing additional area and energy overhead.

\textbf{Novelty in the \newPE design:} \microAdd{The FRI-MAC and TRI-MAC operations allow a single-cycle computation of a recurrent element-wise vector multiplication. A row of \newPE performing FRI-MAC or TRI-MAC operations compute and store the internal-state vector of the Structured-SSMs for S4 and Liquid-S4 layers. \newPE’s support for both real and complex-valued data broadens \acceleratorname’s applicability to a wider range of SSM models.}
\revA{Each \newPE integrates a programmable clock controller within its compute and load buffer units, enabling mode-aware scheduling through clock gating of decoupled preload and compute phases, including a sleep mode that fully disables both units.}

\revCommon{\textbf{Overhead of reconfigurability in \newPE design:} Reconfiguration is achieved via dedicated instructions, altering the operating mode of each PE. 
The hardware overhead to provide this reconfigurability is quantified in Table \ref{tab:pe-type-table} (Row: \newPE\(\ddag\)), showing a 1.3–1.7\(\times\) increase in area and up to a 1.1\(\times\) increase in power for the PEs. 
The additional logic for reconfiguration causes longer critical paths, reducing the maximum operating frequency (\(f_{max}\)) by 5\(\%\).}

\subsection{\dataflowname dataflow}
\label{sec:ProDF}
\microAdd{A unified accelerator is essential for efficiently supporting diverse workloads. Optimal dataflows vary depending on the model type—such as S4, Liquid-S4 and DNN—and are also influenced by the dimensions of the weight, input, or output matrices.}
Previous approaches to efficiently support multiple dataflows have either relied on combining accelerators with distinct dataflows, as demonstrated by Xu et al~\cite{HeSA}, or on modifying interconnection routes between PEs using programmable controls as shown by Tong et al~\cite{FEATHER} and Chen et al~\cite{EYERISS}. 
Both strategies demand significant hardware design complexity. For example, the proposal by Tong et al ~\cite{FEATHER} requires the additional reorder reduction switch circuitry, introducing a 15\% area overhead compared to the proposed dataflow. 
Furthermore, this extra circuitry is not logically adjacent to the processing elements (PEs), causing additional delays due to complex routing.

In this paper, we introduce a novel approach to support both multiple general-purpose dataflows and specialized dataflows for S4 and Liquid-S4 models. 
This novel dataflow, termed \dataflowname, programmatically modifies the flow of data within individual PEs while keeping the interconnection circuitry between PEs unchanged. 
\revD{Dataflow for S4 and Liquid-S4 layers during the compute phase can be broken down into the following pipelined stages:}

\revD{\textbf{Layer I - Efficient Scalar-Vector Multiplication}: 
In S4 and Liquid-S4 models, we scale the current input by a vector of coefficients (as illustrated in Layer I of Figure \ref{fig:Hippo_NN_Layer}). 
To efficiently perform this task within a 2D SA, a single row of \newPEs can be used in parallel, enabling a 1-cycle scalar-vector multiplication. 
This requires the \newPEs to be configured for a Banded-Matrix WS MAC operation, as depicted in Figure \ref{fig:LimaMacOperatingModes}c. 
Unlike traditional WS MAC, where inputs and results flow in horizontal and vertical directions, respectively, our method feeds inputs diagonally into the SA and propagates partial results vertically downward to the next step of the Structured-SSM model.}

\revD{This design allows processing of a new input every cycle, irrespective of the vector size \(N\), where \(N\) represents the state-map size of the S4 and Liquid-S4 models. 
By contrast, traditional WS or OS dataflows would require \(N\) clock cycles to complete the operation for each data in the input sequence, making our approach significantly more efficient.}

\revD{\textbf{Layer I - Efficient Recurrent Integration with Diagonal Dataflow}:
Recurrent integration, as illustrated in Layer I of Figure \ref{fig:Hippo_NN_Layer}, can be conceptualized as a diagonal matrix-vector operation, much like the scalar-vector product. 
A single row of \newPEs can be allocated to process this operation in a single cycle. 
For this, the PEs must be configured to operate in either the Fixed Recurrent Integration (FRI) MAC mode for S4 models or the Time-Varying Recurrent Integration (TRI) MAC mode for Liquid-S4 models, as depicted in Figures \ref{fig:LimaMacOperatingModes}(a) and \ref{fig:LimaMacOperatingModes}(b). 
These modes enable input data to flow vertically through the array while results propagate diagonally to the next step of the Structured-SSM layer.}

\revD{This innovative dataflow ensures seamless transfer of results to subsequent Structured-SSM layers, producing the entire output vector from Layer I simultaneously.
In contrast, traditional dataflows would require a full 2D PE array, resulting in increased latency and energy consumption.}



\revD{\textbf{Optimized Layer II - Matrix Multiplication for S4s}:
Layer II matrix multiplication in S4 and Liquid-S4 models, as illustrated in Figure \ref{fig:Hippo_NN_Layer}, is heavily influenced by the lengths of the input and output sequences. 
On traditional SAs, while WS dataflows are well suited for such operations, they require the input to be staggered for lower latency. 
The novel diagonal dataflow offers a more efficient alternative by enabling simultaneous feeding of the results from Layer-I into Layer-II without staggering. 
This approach propagates, the unaltered Layer-II input, diagonally through the SA while partial results are propagated vertically downward, achieving a streamlined computation. 
Additionally, diagonal dataflow is particularly advantageous for sparse weight matrices, such as banded matrices, making evaluation highly effective for these specialized models.}

\textbf{Unified SA computation for Layer-I and Layer-II of S4s}:
The compact arrangement of processing elements (PEs) optimized for both Layer I and Layer II computations enables a unified approach within a single 2D SA structure. In a traditional SA, scalar-vector products, recurrent integration, and matrix multiplication are executed as separate operations, incurring additional overhead from intermediate read/write operations to on-chip SRAM when transferring outputs between steps.
The proposed dataflow eliminates this overhead by allowing all three operations to be seamlessly performed across consecutive rows of the 2D SA. This integration significantly enhances performance, resource utilization, and energy efficiency, streamlining the computation of S4 and Liquid-S4 models.

\revD{Following are the \textbf{key innovations in \dataflowname} compared to Eyeriss~\cite{EYERISS} and FusedCNN~\cite{FusedCNN}.
The row-specialized pipeline in \acceleratorname is explicitly optimized for temporal and recurrent structure, enabling efficient mapping of SSMs, unlike the homogeneous tiles in Eyeriss and FusedCNN. 
\acceleratorname’s programmable MACs are tailored for the algebraic diversity of structured SSMs, while conventional accelerators hardwire general-purpose MACs. 
Fully in-situ SSM execution in \acceleratorname allows recurrent models to be evaluated with zero intermediate storage overhead, which neither Eyeriss nor FusedCNN can achieve due to their layer-based compute flow.}

\begin{figure}[b]
    \centering
    \vspace{-0.15in}
    \includegraphics[width=0.48\textwidth]{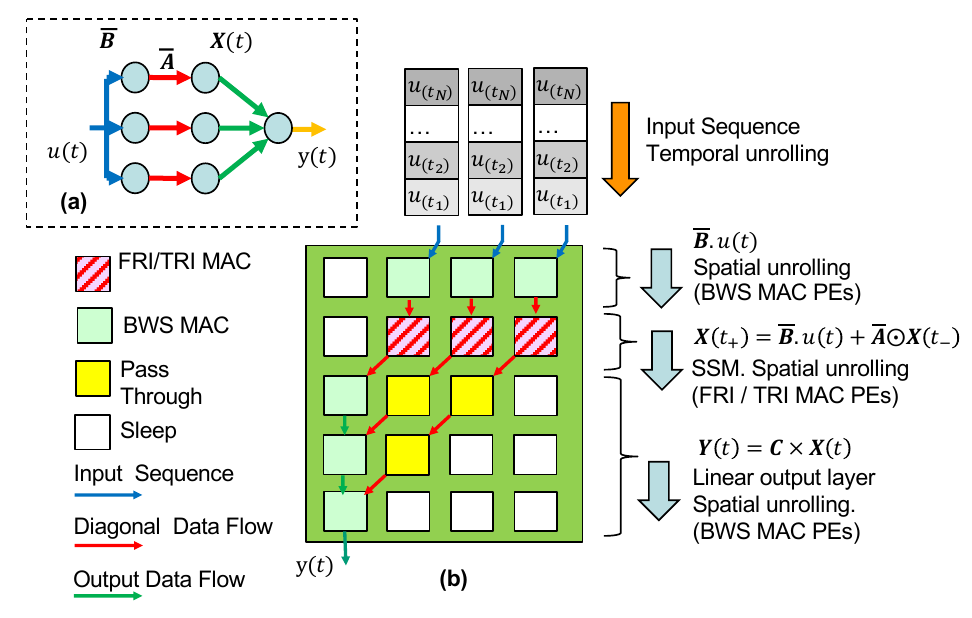}
    \vspace{-0.15in}
    \caption{(a) Example S4 network with state-map size \(N=3\) and number of heads, \(H=1\).(b) The layout of PE modes for a unified computation of Layer-I and Layer-II of S4 models. Data flow of the continuous S4 layer involves two phases. The pre-load data flow loads weights and control bits that program the operating mode of each PE. The compute phase reads the input data sequence and scales it by \(\mathbf{\overline{B}}\) in the first row. The result is passed to the second row that does continuous leaky-time integral with HiPPO matrix \(\mathbf{\overline{A}}\), taking advantage of the diagonal \(\mathbf{A}\) matrix. The remaining rows perform a linear transformation on the state-map layer and transfer the resulting sequence to the on-chip SRAM.
    }
    \label{fig:s4_mode}
\end{figure}

\subsection{Mapping S4 to \acceleratorname}
In this section, we present how we leverage the innovations in \newPE microarchitecture and \dataflowname to efficiently map S4 to \acceleratorname.

\subsubsection{Setting PE Modes}
As outlined in the section \ref{BG_SSM}, S4 models process each element of the input sequence to generate \(H\) continuous output sequences through Scalar-Vector Multiplication, Recurrent Integration, and Matrix Multiplication.
Figure \ref{fig:s4_mode}b illustrates the various PE operating modes in the 2-D SA for evaluating an S4 network with a state-map size of \(N=3\) and a single output sequence (\(H=1\)). 
Each element of the input sequence is fed to the first row of PEs at every clock cycle. 
The PEs in the first row are configured to perform scalar-vector multiplication using BWS MAC modes. 
The input sequence is applied diagonally to the first row of PEs, which propagates scaled values downward to the next row. 
The second row executes recurrent integration, updating the state-map vector corresponding to the input sequence. 
The remaining PE rows perform matrix multiplication, configured to use \emph{Pass Through}, \emph{BWS MAC} or \emph{Sleep} modes as needed. 

\subsubsection{Dataflow}
Next we describe the data flow for S4 network as shown in Figure \ref{fig:s4_mode}b. 
For S4 layers with state-map size of \(N\), the SA should be of size \((N+2)\times (N+1)\).
During the \emph{Pre-Load Phase}, a vector of $(N+2)$ values, consisting of weights and PE control signals, is read from the on-chip weight SRAM and loaded into the first column of $(N+2)$ PEs. 
Over the subsequent $(N+1)$ cycles, all the \((N+2)\times (N+1)\) PE tiles are preloaded with these weights and control values.
In the \emph{Compute Phase}, (Figure \ref{fig:s4_mode}b), the input sequence is fed from the on-chip input/output SRAM to the first row of PEs. 
The first data element of the output sequence becomes available after $(N+2)$ cycles, and subsequent outputs are generated every clock cycle. 
The evaluation order is illustrated in Figure \ref{fig:s4_dataflow}a. 
Once processing of a single input sequence is complete, the SA resets the buffered state-map and partial results of all PEs, while retaining the buffered weights. 
However, both weights and partial results must be reset between input batches.

\subsection{Mapping Liquid-S4 to \acceleratorname}
\begin{figure*}[t]
    \centering
    \includegraphics[width=0.8\textwidth]{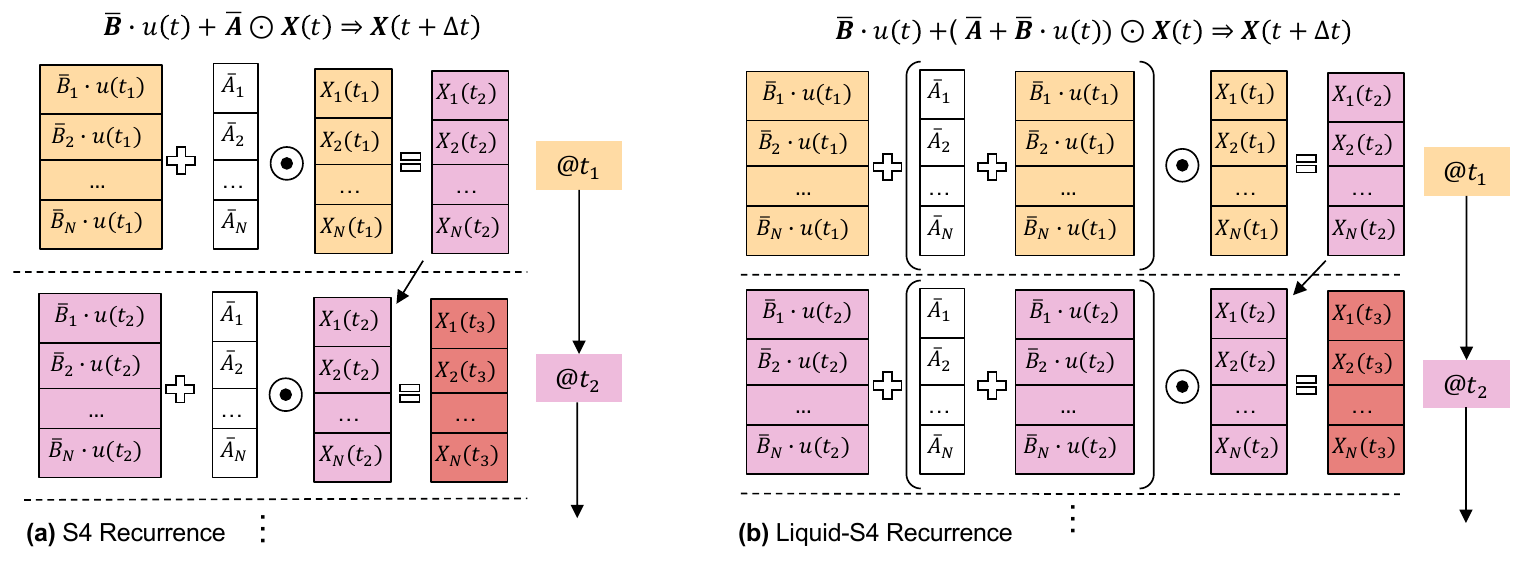}
    \vspace{-0.25in}
    \caption{(a) S4 recurrence includes addition of linearily scaled input token \(\mathbf{\overline{B}}\cdot u(t)\) and element-wise multiplication of time-invariant coefficient \(\mathbf{\overline{A}}\) to the previous hidden state \(\mathbf{X}(t)\), at every time unit. (b) In Liquid-S4 recurrence the linearily scaled input token is also added to the time-invariant coefficient \(\mathbf{\overline{A}}\), resulting in a time-varying coefficient to the hidden state. In both cases the output sequence takes \(N+2\) cycles to compute the first token, thereafter the output element is available at every clock cycle, where \(N\) is the hidden state-map size.}
    \label{fig:s4_dataflow}
    \vspace{-0.15in}
\end{figure*}

As explained in Section \ref{BG_LIQ_SSM}, the Liquid-S4 network shares computational similarities with the S4 network, and all \acceleratorname innovations for the S4 network are also relevant to the Liquid-S4. 
One key difference is in the recurrent integration step. 
Liquid-S4 introduces a time-varying, input-dependent coefficient, which requires the recurrent integration PEs to operate in TRI-MAC mode.
The dataflow of the Liquid-S4 network largely mirrors that of the S4 network (as illustrated in Figure \ref{fig:s4_mode}b). 
The main difference is that the second-row PEs are preconfigured to execute the TRI-MAC operation. 
Figure \ref{fig:s4_dataflow}b highlights the specific operations of the Liquid-S4 network.

\subsection{Mapping Other Networks to \acceleratorname}\label{DNN_EpochCore}
\subsubsection{CNN/RNN/Transformers}
\acceleratorname can be configured to execute a variety of neural network operations, such as GEMM for CNN, RNN and Transformers, by supporting common dataflows like WS, IS and OS. 
Figure \ref{fig:OS_DataFlow} demonstrates how \acceleratorname can be utilized for a DNN operation using the OS dataflow.
In this configuration, all \newPE are preset to perform the Traditional OS (TOS) MAC operation (Figure \ref{fig:LimaMacOperatingModes}d) by proper selection of multiplexers (M2, M3, M4 and M6). 
The inputs and weight vectors are read every cycle and fed to the SA. 
The input data flows in the West-to-East direction, and the weights flow in the North-to-South direction. 
At the end of the GEMM operation, the entire output matrix is read out. 
The IS and WS dataflows can be similarly implemented. 
Transformers predominantly use GEMM operations as well, and their applications are complementary to those of SSMs.
So, transformers can be executed on \acceleratorname.

\subsection{\acceleratorname for other SSMs}\label{MAMBA_S4}
Patro et. al \cite{ssm_survey_2024} provide a taxonomy of recently proposed SSMs, categorizing them based on their structural, gated and recurrent characteristics. 
Within the structured category, models such as H3 \cite{H3} implement two layers of structured SSMs: one performing a shift operation, and another utilizing a diagonal state matrix. 
The shift-SSM layer in such models can be executed within \acceleratorname's SA as a single row of \newPE by including additional sleep PE at the beginning of the row and shifting the row to the right. SSMs in the gated category improve contextualization, as seen in GSS \cite{SSM4}. 
Mapping GSS models onto \acceleratorname involves evaluating the associated linear layers by appropriately setting the mode of \newPE, followed by the evaluation of an on-chip non-linear activation function to implement gating. 
\revCommon{The Mamba model~\cite{gu2023mamba, li2025mamba}, shares key characteristics with Liquid-S4~\cite{LIQ_SSM1, LIQ_SSM2}, including the use of input-dependent time-varying state-matrix.
Mamba also builds upon the framework introduced in H3 \cite{H3} and GSS \cite{SSM4}. 
In Mamba, the discretized coefficients \(\mathbf{\overline{A}}\) and \(\mathbf{\overline{B}}\) vary with input, while the discretization step \(\triangle t\) effectively serves as a gating mechanism. Supporting Mamba models on
\acceleratorname requires pre-loading the inputs to compute input-dependent discretized coefficients on the host. 
These updated coefficients must be loaded onto \acceleratorname as weight matrices, introducing additional overhead for input processing and data transfer. 
This overhead is roughly \(N\times\) the performance cost of S4, where \(N\) is the state-map size, making \acceleratorname highly inefficient. 
To efficiently support Mamba, two modifications to \acceleratorname are required. (1) Support variable discretization (\(\Delta t\)) via additional \newPE rows, (2) new MAC operations in \newPE to support gated element-wise multiplication. 
These modifications to \acceleratorname, to make it generic, are part of our future work.
}


\subsection{Other Components}\label{OTHER}
A complete neural network model involves additional mathematical operations at the output of each layer. \emph{\acceleratorname} integrates on-chip specialized circuitry to support a range of non-linear activation functions (such as SiLU, ReLU, Sigmoid and TanH) and layer normalization, as shown in the \acceleratorname micro-architecture in Figure \ref{fig:EpochCoreMicroArch}b. 
These operations are implemented as custom digital units.
\vspace{-0.2in}

\revB{\subsection{\acceleratorname for training}\label{TRAINNG}
The inference operations are typically a subset of the training operations. 
When training S4 and Liquid-S4 on GPUs, these inference operations typically account for 10-30\(\%\) of total latency. 
The weight update calculations for SSM layers is computationally similar to those of conventional layers like DNNs, relying on GEMM operations to compute gradients. 
\acceleratorname can also be used for training by executing GEMM operations for weight updates, though it consumes a \(1.3\times\) higher energy consumption compared to TPUs. 
Like all other accelerators, the non-linear operations are executed in separate dedicated units.
}

%% file: evaluation.tex
\vspace{-0.1in}
\section{Evaluation}\label{EVAL}
In this section, we compare \acceleratorname accelerator against other SoTA accelerators. The evaluation results apply to both inference and training. Training of neural models requires a forward pass and a backward pass, while inference involves a forward pass. The backward pass includes two GEMM operations for weight updates. In this paper, we evaluate \acceleratorname for inference i.e. the forward pass using 32-bit fixed-point precision. Improving forward pass performance benefits training as well.

\subsection{Methodology}
We designed \newPE at the RTL level and synthesized it using Free-PDK45~\cite{FreePDK45}, along with NanGate's standard cell library. 
For synthesis, we utilized Cadence's Genus\texttrademark{} tool \cite{Genus} to meet a target clock frequency of 700MHz. We created test benches and simulated 1 million cycles using Cadence's NC-Sim\texttrademark{}, generating a  
Value Change Dump (VCD) file. This file was then fed back into Genus\texttrademark{} to evaluate the quality of results (QoR) using power, area, and performance metrics. 
To evaluate the \newPE, we compared it against other PEs from the literature. We accounted for differences in technology nodes by applying appropriate technology scaling techniques \cite{STILLMAKER}.

\begin{figure}[b]
    \centering
    \vspace{-0.05in}
    \includegraphics[width=0.5\textwidth]{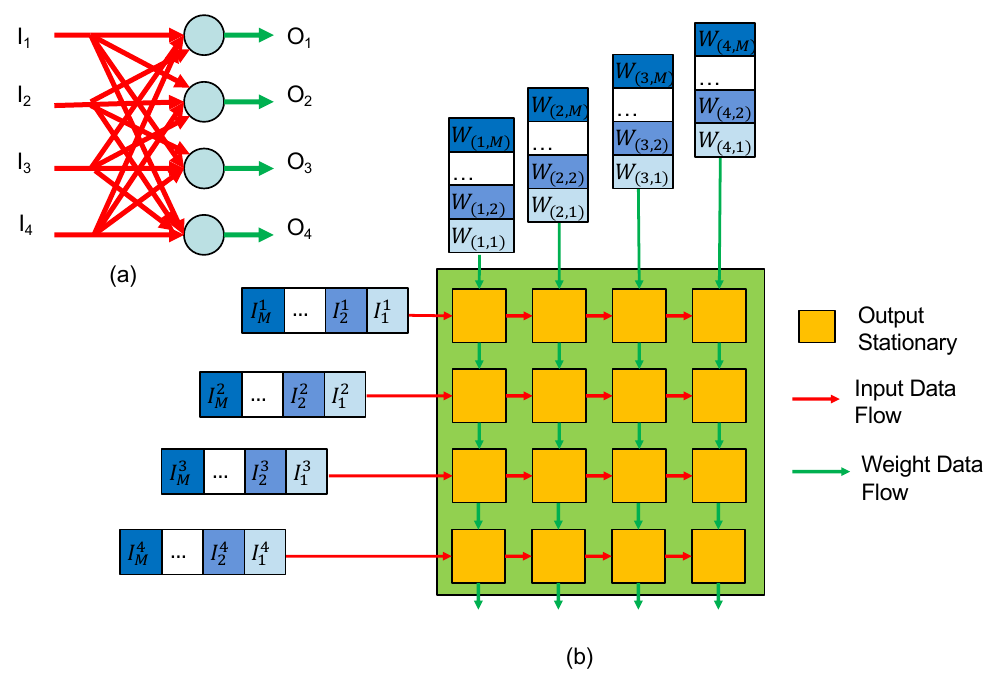}
    \vspace{-0.3in}
    \caption{The \acceleratorname can be programmed to compute regular GEMM. An example usage for output-stationary (OS) GEMM data flow within \acceleratorname is shown.}
    \label{fig:OS_DataFlow}
\end{figure}

To evaluate \acceleratorname, we developed a custom cycle simulator based on ScaleSim \cite{ScaleSim} to estimate key metrics such as cycle count, throughput, and energy consumption for various configurations. 
For evaluating S4 and Liquid-S4 based models, we used model parameters cited by Gu et al ~\cite{SSM2}. 
Our analysis is based on LRA datasets (Table \ref{tab:lra-datasets}), where S4-based neural models remain SoTA. Additionally, we created a memory bandwidth simulator to measure continuous bandwidth requirements of \acceleratorname and other accelerators under LRA workloads. 
The memory access time, using the same technology i.e. 45 nm as the PE design, was determined using a CACTI-based memory compiler~\cite{CACTI}.

\begin{table}[]
\caption{Comparison of PEs. Traditional-PE, STPU-PE and four types of \newPE for FixedPoint32 and Int8 operation}
\vspace{-0.1in}
\label{tab:pe-type-table}
\scalebox{0.9}{
\begin{tabular}{ccccc}
\hline
\rowcolor[HTML]{C0C0C0} 
\cellcolor[HTML]{C0C0C0} & \multicolumn{2}{c}{\cellcolor[HTML]{C0C0C0}\textbf{FixedPoint32}} & \multicolumn{2}{c}{\cellcolor[HTML]{C0C0C0}\textbf{Int8}} \\ \cline{2-5} 
\rowcolor[HTML]{C0C0C0} 
\multirow{-2}{*}{\cellcolor[HTML]{C0C0C0}\textbf{\begin{tabular}[c]{@{}c@{}}PE \\ Design\end{tabular}}} & \textbf{\begin{tabular}[c]{@{}c@{}}Area \\  (\(\mu m^2\))\end{tabular}} & \textbf{\begin{tabular}[c]{@{}c@{}}Power \\ (\(mW\))\end{tabular}} & \textbf{\begin{tabular}[c]{@{}c@{}}Area\\ (\(\mu m^2\))\end{tabular}} & \textbf{\begin{tabular}[c]{@{}c@{}}Power\\ (\(mW\))\end{tabular}} \\ \hline
Trad-PE \cite{AccelANN2} &3,527 (1.0X)&7.4 (1.0X)&474 (1.0X)&0.84 (1.0X)\\ \hline
STPU-PE \cite{STPU} &17,487 (4.9X)&9.0 (1.2X)&1,690 (3.6X)&1.07 (1.3X)\\ \hline
\newPE\(\dag\) &4,606 (1.3X)&5.9 (0.8X)&493 (1.0X)&0.56 (0.6X)\\ \hline
\newPE\(\ddag\) &6,021 (1.7X)&8.4 (1.1X)&611 (1.3X)&0.74 (0.8X)\\ \hline
\newPE&7,221 (2.0X)&11.5 (1.6X)&677 (1.4X)&0.89 (1.0X)\\ \hline
\newPE (CP16) &8,198 (2.3X)&12.6 (1.7X)&-&-\\ \hline
\end{tabular}}
\vspace{-0.1in}
\end{table}


\subsection{\newPE vs other PEs}
In Table \ref{tab:pe-type-table}, we compare the power consumption and area of the \newPE against other published PE designs, using the traditional PE from TPUs \cite{AccelANN2} as reference. 
The \newPE shows a 1.4-2\(\times\) increase in area, and power consumption increases by 1-1.6\(\times\) across the 8-bit and 32-bit versions.

Sparse-TPU (STPU)~\cite{STPU} was designed to handle GEMM with large sparsity, featuring a PE that supports special modes. When comparing \newPE to STPU-PE, \newPE demonstrates a smaller area footprint for both the 8-bit and 32-bit versions. 
\newPE is more energy-efficient than STPU-PE in the 8-bit version but consumes more power in the 32-bit version. 

We also analyzed the area and power impacts in the development of \newPE design compared to traditional PE. 
The \newPE\(\dag\) design consists of only the \emph{green} logic (Figure \ref{fig:LimaMacOperatingModes}), which supports only the traditional MAC operations. 
The \newPE \(\ddag\) design adds additional circuitry, incorporating the \emph{yellow} logic (Figure \ref{fig:LimaMacOperatingModes}), to support FRI and BWS MAC operations for S4 models. 
The fully-loaded \newPE design further includes \emph{blue} logic (Figure \ref{fig:LimaMacOperatingModes}) to support TRI MAC operations for Liquid-S4. 
The \newPE (CP16), supports 16-bit real and imaginary part of a 32-bit data, and includes the unified multiplier compute unit to support real and complex valued data. 
The area and power overhead due to the addition of supporting both 32-bit real and 16-bit complex data types is not a significant increase.
The progression of area and power increase can be seen as more logic is introduced.


The 20-40\% reduction in power at the cost of a 30\% increase in area is presented in Table \ref{tab:pe-type-table}, comparing rows 1 (Trad-PE without clock-gating) and 3 (\newPE with Trad-MAC and with clock-gating).
Power consumption for various \newPE operating modes was measured and summarized in Table \ref{tab:lima_pe_mode_power}. 
The Sleep and Pass-Through modes provide substantial power savings, especially when higher bit-precision is used.

\begin{table}[t]
\begin{center}
\caption{Power across \newPE Modes}
\vspace{-0.1in}
\label{tab:lima_pe_mode_power}
\begin{tabular}{lll}
\hline
\rowcolor[HTML]{C0C0C0} 
\textbf{\newPE Mode} & \textbf{\begin{tabular}[c]{@{}l@{}}FixedPoint32 \\ Power (\(mW\))\end{tabular}} & \textbf{\begin{tabular}[c]{@{}l@{}}Int8\\ Power (\(mW\))\end{tabular}} \\ \hline
Sleep & 3.8 & 0.54 \\ \hline
Pass-Through & 6.7 & 0.73 \\ \hline
Accumulation & 11.5 & 0.89 \\ \hline
\end{tabular}
\vspace{-0.05in}
\end{center}

\end{table}






\subsection{\acceleratorname vs Systolic Arrays}\label{EC_VS_SA}



S4 and Liquid-S4 models involve specialized operations, such as scalar-vector product and recurrent integrations that have sparsity and so they are not well suited to be run on regular SA. 
To evaluate \acceleratorname for these models, we compare its performance against other SA-based architectures such as the Sparse-TPU \cite{STPU} and Boriakoff's \cite{FFT_SA} FFT-based SA accelerator. 
The datasets used in this evaluation were selected based on software evaluations of S4 and Liquid-S4 done by Gu et al~\cite{SSM2}, \revD{which spans vision, audio, and other challenging LRA tasks commonly used to benchmark state-of-the-art model accuracy~\cite{li2025mamba}.} For our experiments, we focused on a single Liquid-S4 layer with a state-map size of \(N=64\), and varied sequence lengths, \(T\) as shown in Table \ref{tab:lra-datasets}.  

\begin{figure}[t]
    \centering
    \includegraphics[width=0.48\textwidth]{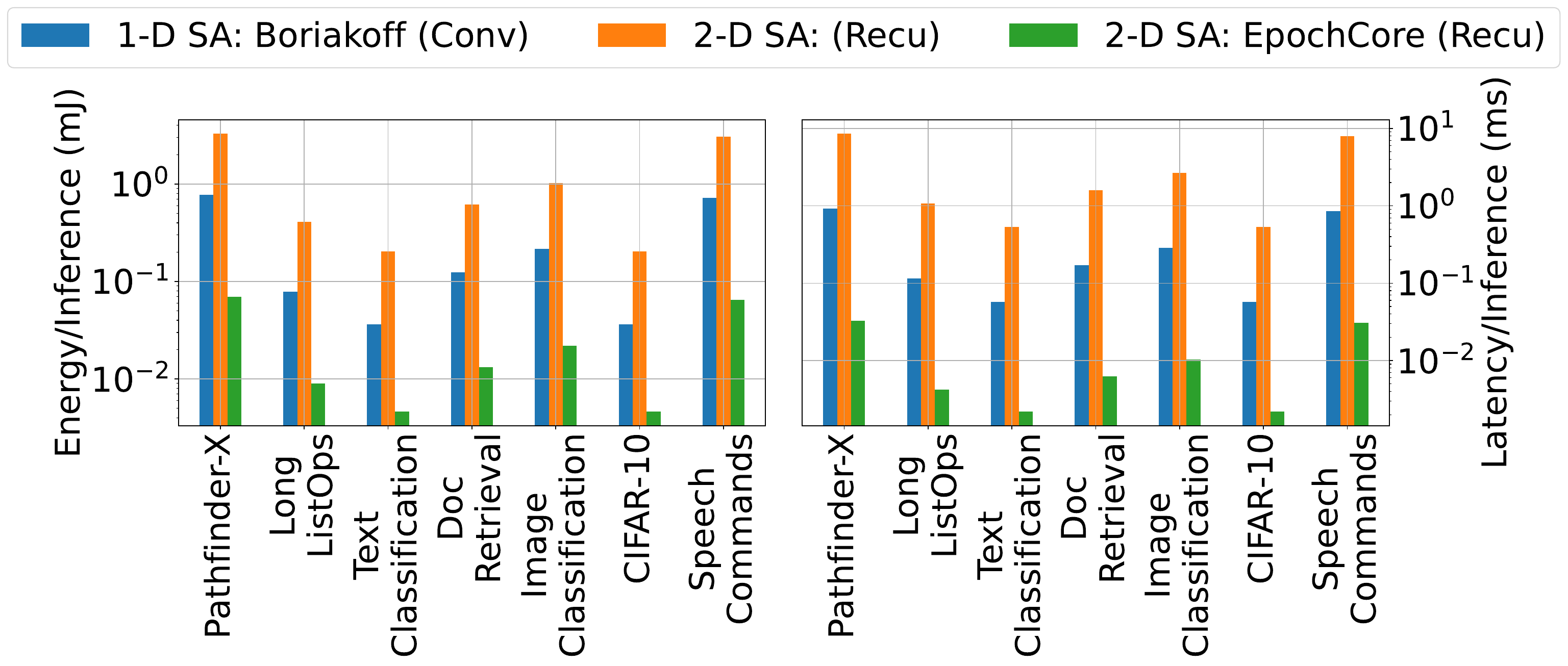}
    \vspace{-0.2in}
    \caption{Energy/Inference and Latency/Inference for various LRA datasets shown in Table \ref{tab:lra-datasets} for Liquid-S4 model when using Boriakoff-based 1-D SA \cite{FFT_SA}, Sparse 2D SA, and \acceleratorname.}
    \label{fig:edp_ssm_exp}
    \vspace{-0.05in}
\end{figure}

\subsubsection{Energy and Latency Comparison}\label{EN_COMPARE}
The energy and latency per inference for a set of LRA datasets using Boriakoff's, Sparse-SA and \acceleratorname are show in Figure ~\ref{fig:edp_ssm_exp}. On average, \acceleratorname achieves \(250\times\) lower latency than Sparse-SA and \(45\times\) lower latency compared to Boriakoff's SA accelerator. Also, the energy consumption shows consistent improvement by \(25\times\) and \(10\times\) on an average over  Sparse-SA and Boriakoff's SA accelerator, respectively. The improvements in \acceleratorname's performance over Sparse-SA are attributed to its specialized dataflow, which is tailored for handling S4 and Liquid-S4 operations. Additionally, \acceleratorname outperforms Boriakoff's SA accelerator by avoiding simultaneous movement of long-sequence kernel weights and input elements, a factor that contributes to higher latency and energy use in Boriakoff's design.

\begin{figure}[t]
    \centering
    \includegraphics[width=0.46\textwidth]{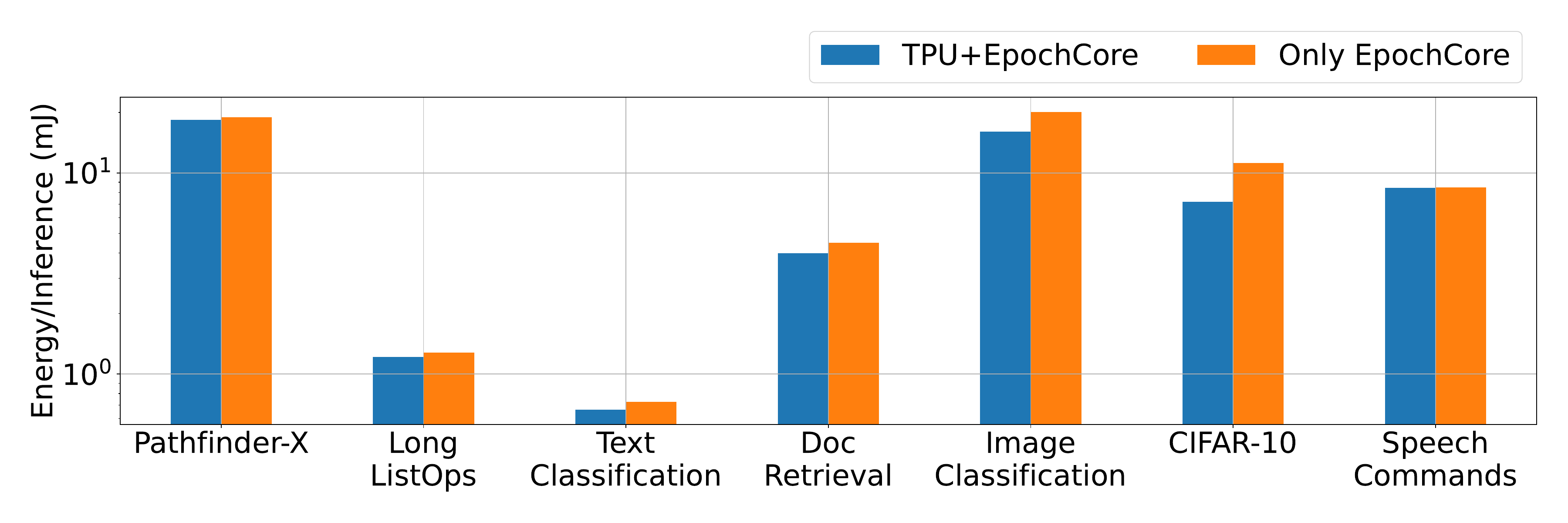}
    \vspace{-0.05in}
    \caption{\revE{Energy/Inference for executing Liquid-S4 and DNN layers for various LRA datasets shown in Table \ref{tab:lra-datasets} when using TPU + \acceleratorname and only \acceleratorname accelerators. The overhead of \acceleratorname on DNN layers is negligible.}}
    \label{fig:soc_energy}
    \vspace{-0.15in}
\end{figure}

\begin{figure}[!ht]
    \centering
    \vspace{-0.05in}
    \includegraphics[width=0.48\textwidth]{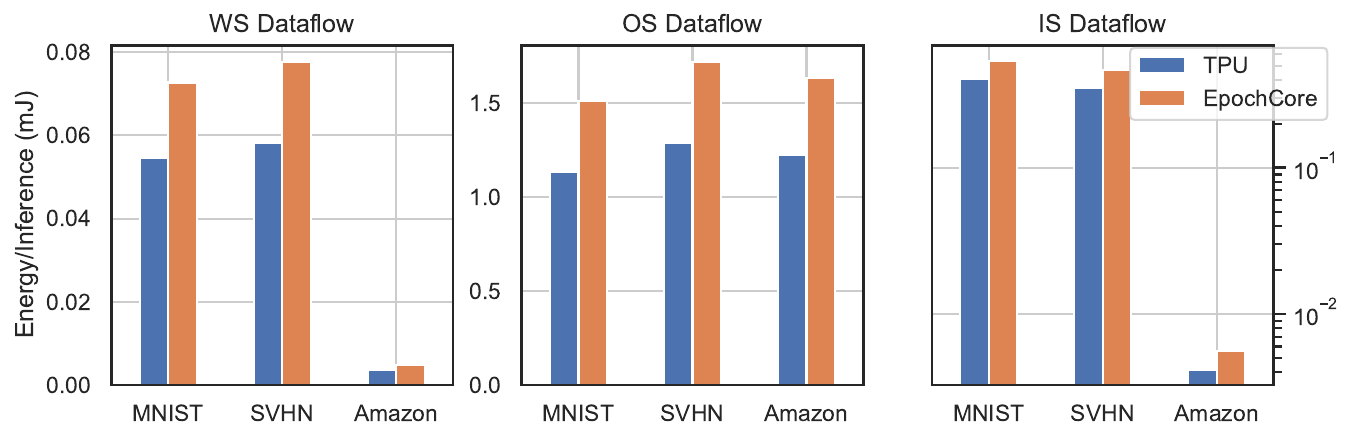}
    \caption{Energy/Inference for TPU and \acceleratorname for WS, OS and IS dataflows for various DNN datasets.}
    \label{fig:edp_dnn_exp}
\end{figure}

\revB{For GEMM with WS, IS and OS dataflows, \acceleratorname uses the same cycles and PE units as TPU-SA. We calculated the energy per inference with 32-bit data for WS, OS and IS dataflows using ScaleSim~\cite{ScaleSim} for the DNN datasets in Table~\ref{tab:dnn_experiments}, and power values from Table ~\ref{tab:lima_pe_mode_power}}. 
\revD{The experiments were selected to represent deeply layered workloads with large input dimensions.}
Figure~\ref{fig:edp_dnn_exp} shows a \(30\%\) increase in energy per inference across DNNs. 
\revE{Compared to TPU-SA, evaluating GEMM with \acceleratorname has a 2\(\times\) area penalty, 1.3\(\times\) energy penalty and 5\(\%\) increased latency. 
The cost for evaluating the non-linear layer using on-chip components of \acceleratorname is relatively minimal compared to other operations and is comparable to evaluating them on the host. 
Figure~\ref{fig:soc_energy} compares the combined Energy/Inference of evaluating both Liquid-S4 and DNN layers for various LRA datasets. 
The evaluation was done emulating a `TPU + \acceleratorname' system where DNN layers were evaluated on TPU and Liquid-S4 layers on \acceleratorname, as well as an \acceleratorname only system, where both DNN and Liquid-S4 layers were evaluated on \acceleratorname. 
The homogeneous system consumes 30\% higher energy.}

Despite higher energy costs for DNN layers, \acceleratorname’s efficiency and throughput gains in Liquid-S4 layer inference (Figure \ref{fig:edp_ssm_exp}) make it a better choice for real-world applications.

\subsection{Full model evaluation of \acceleratorname}

\begin{figure}[t]
    \centering
    \includegraphics[width=0.48\textwidth]{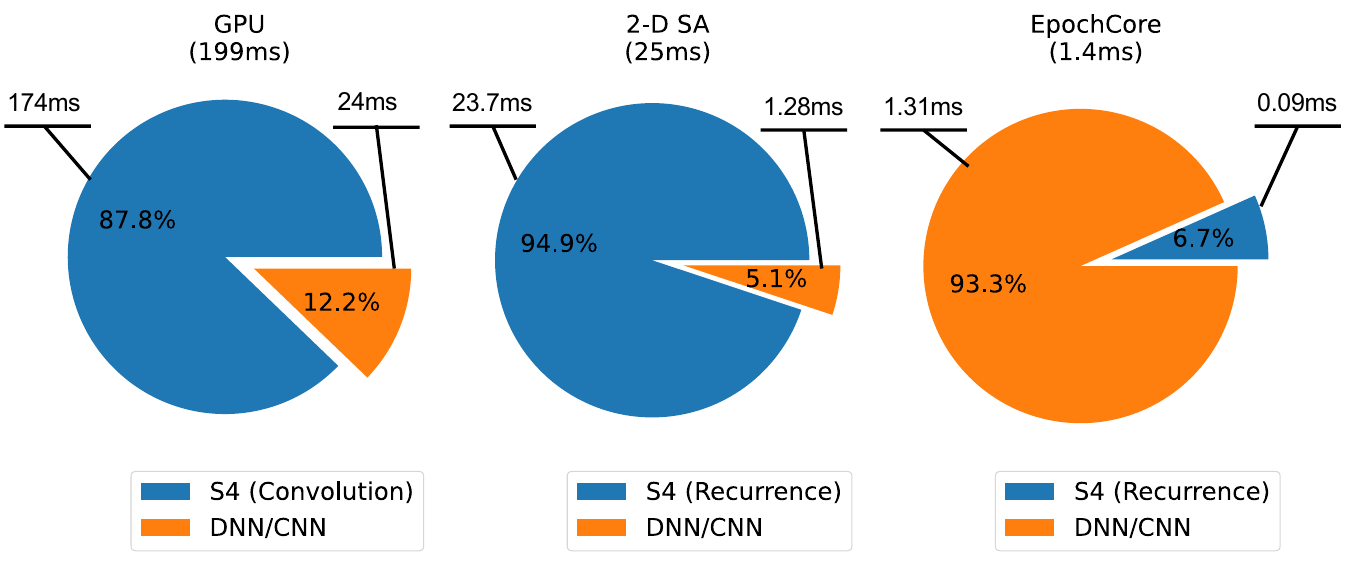}
    \vspace{-0.05in}
    \caption{\revF{Latency breakdown of S4 based models for CIFAR-10 dataset with sequence length, T=64K, using GPU, Sparse-SA and \acceleratorname accelerators.}}
    \label{fig:SSM_PIE_EPOCH}
    \vspace{-0.1in}
\end{figure}

\begin{figure}[!ht]
    \centering
    \vspace{-0.05in}
    \includegraphics[width=0.45\textwidth]{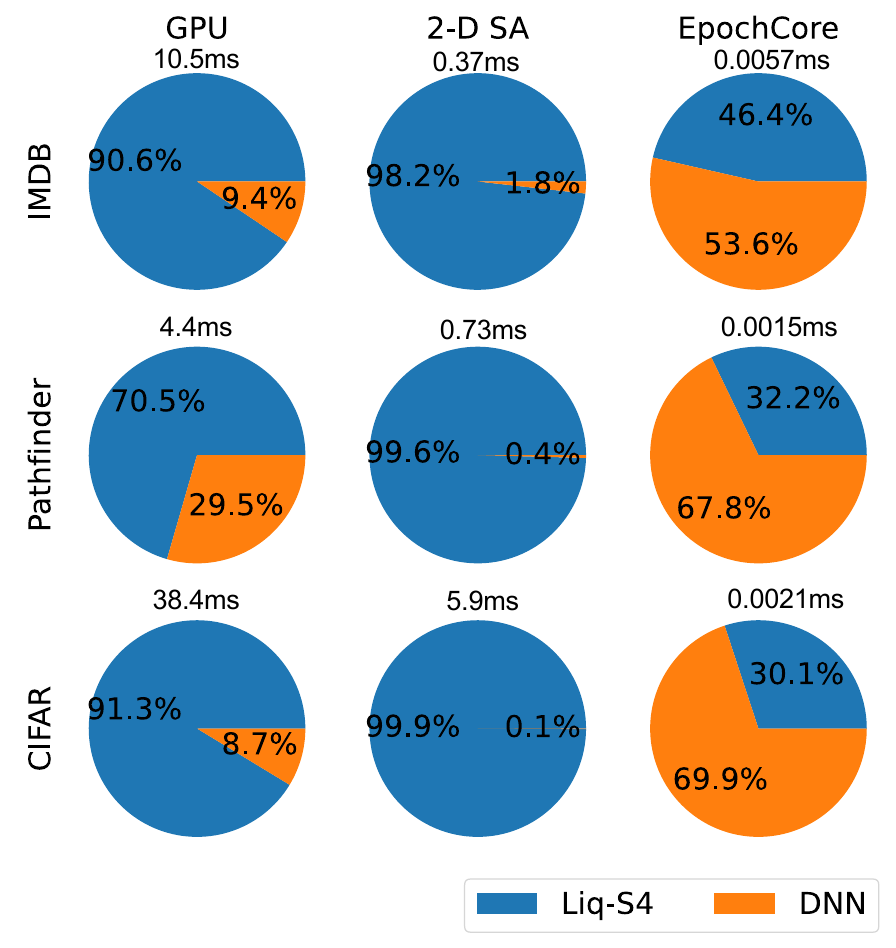}
    \caption{Latency breakdown of inference of Liquid-S4 models for various datasets with hyper parameters used in \cite{LIQ_SSM1}.}
    \label{fig:dataset_gpu}
\end{figure}

\begin{figure}[!ht]
    \centering
    \vspace{-0.05in}
    \includegraphics[width=0.48\textwidth]{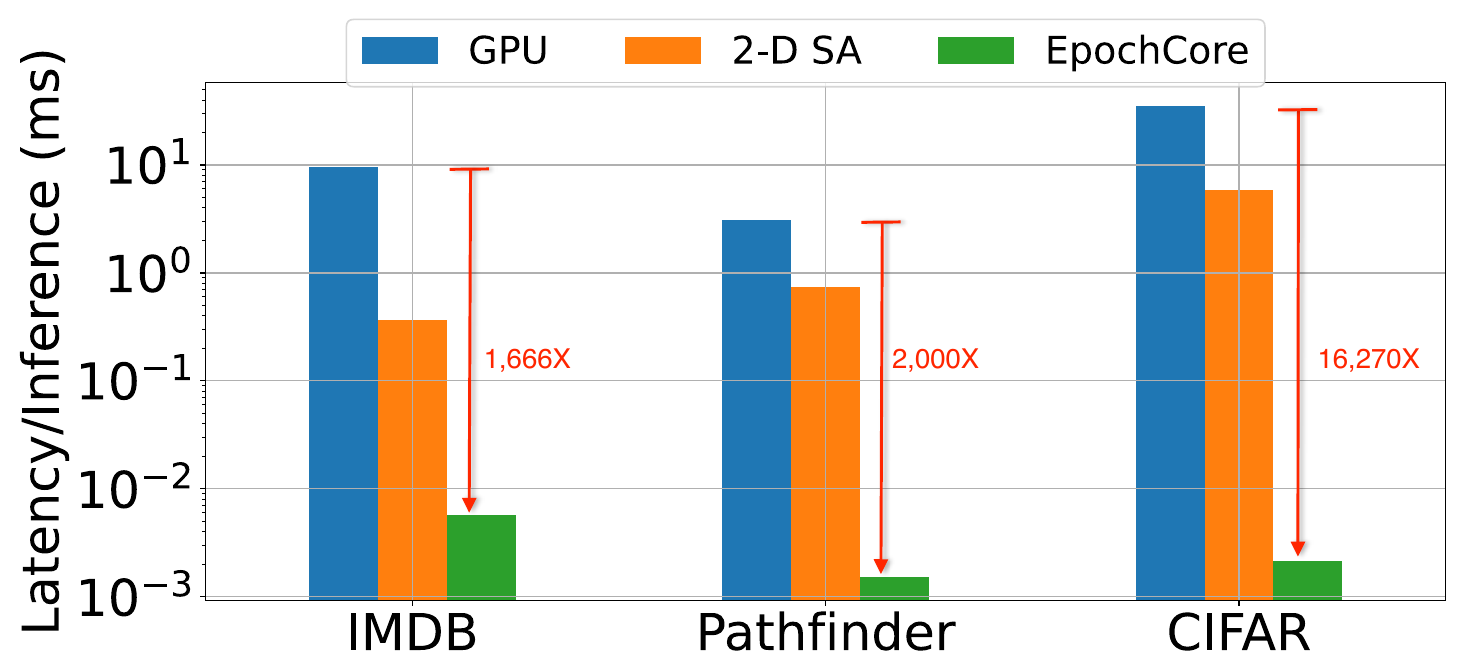}
    \caption{Latency/Inference of Liquid-S4 layers on GPU, TPU and \acceleratorname using LRA datasets. }
    \label{fig:latency_dataset}
\end{figure}

In Figure \ref{fig:SSM_PIE_EPOCH} the latency breakdown between S4 and DNN/SNN layers (as in Figure \ref{fig:SithSsmMultiLayer}) are shown for evaluations on GPU (Nvidia A100), Sparse-SA and \acceleratorname accelerators for the same workload. The \acceleratorname accelerator demonstrates significant improvement in S4 layer latency compared to Sparse-SA. The breakdown also reveals that \acceleratorname significantly reduces the latency of S4 layers from \(\sim\)95\% to \(\sim\)7\%, allowing for a more balanced and efficient execution across different layers. Similarly, latency breakdown on additional LRA datasets using the time-variant Liquid-S4 model is shown in Figure \ref{fig:dataset_gpu} with the corresponding Liquid-S4 model parameters shown in Table \ref{tab:lra-datasets}. The average latency gain for inference of Liquid-S4 layer is shown to be around \(1,666\times\) to \(16,270\times\) as shown in Figure \ref{fig:latency_dataset}. 

\begin{table}[]
\caption{DNN Data Sets and Model Parameters.}
\label{tab:dnn_experiments}
\begin{tabular}{lll}
\hline
\rowcolor[HTML]{C0C0C0} 
\textbf{Dataset} & \textbf{\begin{tabular}[c]{@{}l@{}}DNN Layers \\ Size\end{tabular}} & \textbf{\begin{tabular}[c]{@{}l@{}}Number of \\ Inputs\end{tabular}} \\ \hline
MNIST & {[}784, 500, 400, 300, 100, 10{]} & 4200 \\ \hline
SVHN & {[}1024, 512, 256, 128, 64, 32, 10{]} & 4200 \\ \hline
Amazon Reviews & {[}4364, 16, 8, 1{]} & 3150 \\ \hline
\end{tabular}
\end{table}

\subsection{\acceleratorname vs Other SSM Accelerators}

Table ~\ref{tab:accel-compare} shows up to 3860\(\times\) speedup for H3-SSM models on \acceleratorname, outperforming VGA~\cite{VGA}. 
For Mamba models, \acceleratorname is closely comparable to other accelerators such as FastMamba~\cite{FastMamba} and MARCA~\cite{li2024marca}. \acceleratorname is the first accelerator for Liquid-S4 models.

\subsection{Ablation Study}
In this section we explore the sensitivity of various model parameters to latency and throughput of neural network evaluations using \acceleratorname.

\subsubsection{Impact of Liquid-S4 state-map size on accuracy}\label{EVAL_SSM}
\begin{figure}[b]
    \centering
    \vspace{-0.20in}
    \includegraphics[width=0.5\textwidth]{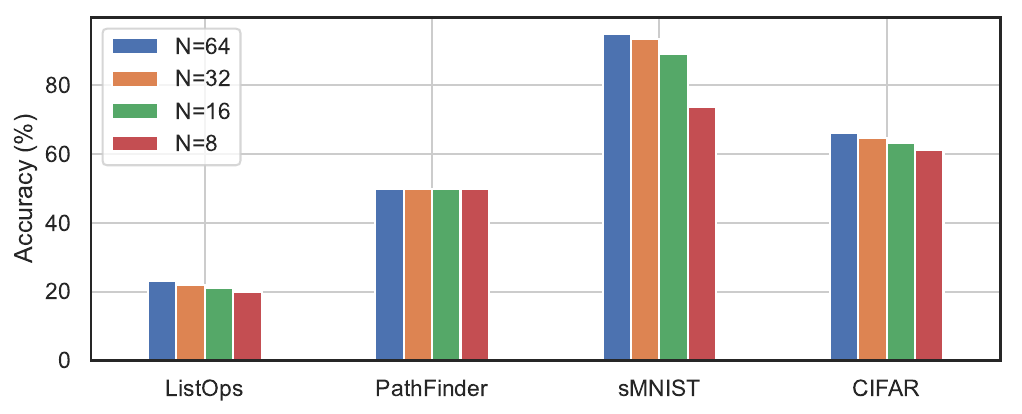}
    \vspace{-0.3in}
    \caption{Accuracy of Liquid-S4 model training with different hidden state-map size \(N\).}
    \label{fig:s4_acc_size}
\end{figure}

In Figure \ref{fig:s4_acc_size} the accuracy impact after ten training iterations is shown for various datasets while varying the hidden state-map size. 
PathFinder requires much larger training iterations to show the sensitivity of state-map size. 
The state map size has no impact on the accuracy of Pathfinder.
For other datasets, choosing the state-map size impacts accuracy. 
Most SoTA models pick the state-map size of \(N=64\).

\subsubsection{On-Chip SRAM Bandwidth and Size Comparison}\label{SRAM_BW}
\begin{figure}[t]
    \centering
    \vspace{-0.05in}
    \includegraphics[width=0.5\textwidth]{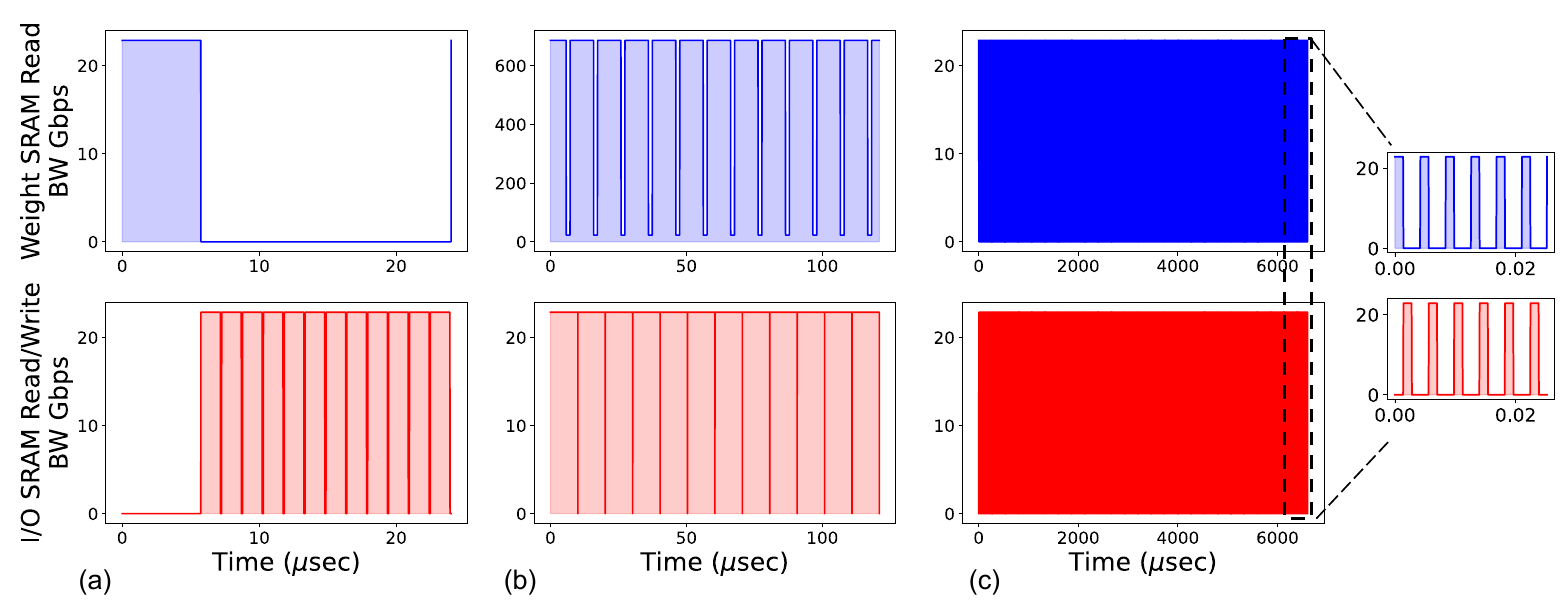}
    \vspace{-0.3in}
    \caption{The Weight and Input/Output on-chip SRAM access bandwidth is shown for (a) \acceleratorname, (b) Boriakoff based 1-D SA, (c) Sparse 2-D SA accelerator, when evaluating an S4 layer with input sequence length, \(T=1024\), state-map size, \(N=64\), batch size, \(B=12\), on-chip SRAM word-size of 32-bits and cycle time of 1.4ns}
    \label{fig:MEM_BW}
\end{figure}

The bandwidth requirements of on-chip SRAM access to weights and input/output data were compared across the three accelerators based on a memory bandwidth simulator that was developed. 
Figure \ref{fig:MEM_BW} shows that 
\acceleratorname requires higher memory bandwidth to load weights prior to the first inference. 
However, for consecutive inferences, the weights remain stationary, freeing up memory bandwidth, which leads to reduced overall bandwidth requirements as batch sizes increase. 
In contrast, Boriakoff's SA architecture requires around \(30\times\) more bandwidth to access on-chip weight SRAM. 
Due to repeated reads of both weight and input data from the on-chip SRAM during and between inferences, the Sparse-SA accelerator requires twice the bandwidth compared to \acceleratorname. 
This is because \acceleratorname minimizes bandwidth usage by keeping weights stationary after the initial load, while Sparse-SA incurs higher bandwidth demands through repeated memory access.

\begin{figure}[b]
    \centering
\includegraphics[width=0.49\textwidth]{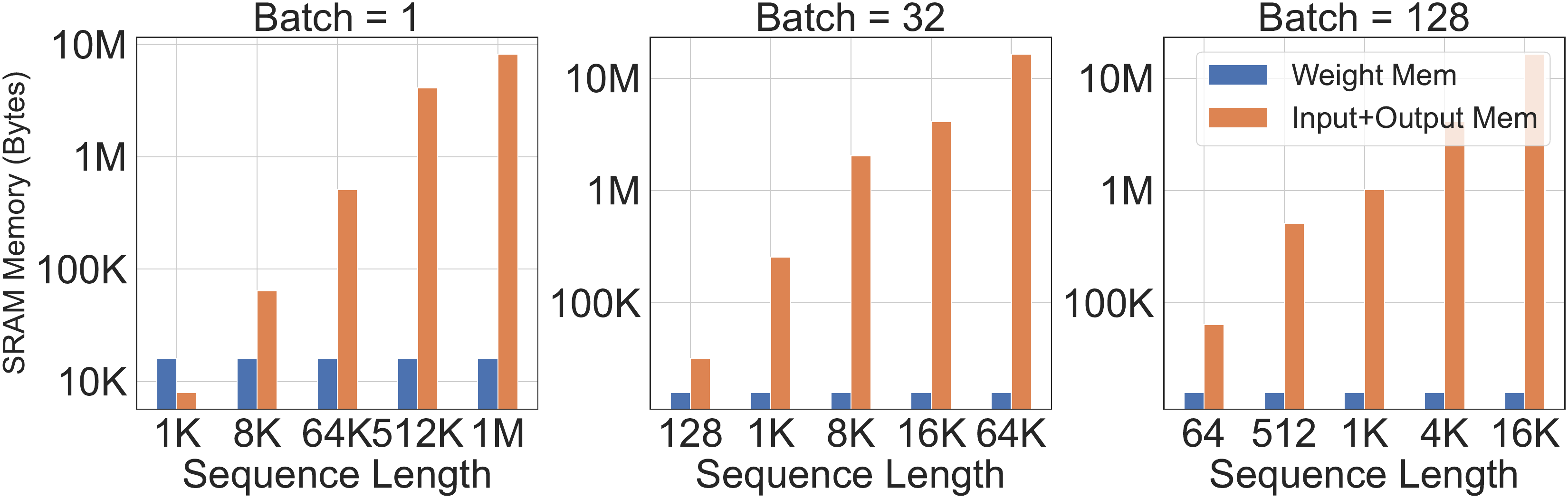}
    \vspace{-0.25in}
    \caption{The Weight and Input/Output SRAM memory sizes for various sequence lengths of S4 and Liq-S4 inputs are analyzed for different input batch sizes. The \acceleratorname with SA size of \(64\times64\) and 32-bit precision data is utilized.}
    \label{fig:MEM_SIZE_SSM}
\end{figure}

The maximum on-chip SRAM size for S4, Liquid-S4 based workloads is determined by the input data sequence length and input batch size. As shown in Figure \ref{fig:MEM_SIZE_SSM}, an input sequence of 1-million, typical for audio and other LRA datasets, can fit within 10MB SRAM with 1-batch input, in a \(64\times64\) SA with 32-bit data. Shorter sequence lengths allow for larger batch sizes within the same SRAM capacity.
\revD{The key observations are: (1) Input/Output SRAMs dominate overall memory usage, and are 2-3 orders of magnitude larger than Weight SRAM; (2) the required Weight SRAM size does not depend on the sequence length; and (3) given a maximum of 10MB SRAM, the figure shows the feasible combinations of batch sizes and sequence lengths (for example a batch size of 32 can support sequence length upto 64K) within a single SSM layer that can be supported at a time by \acceleratorname.}

\begin{table}[]
\vspace{-0.15in}
\begin{center}
\caption{Performance Speed up over GPU}
\label{tab:accel-compare}
\begin{tabular}{llll}
\hline
\rowcolor[HTML]{C0C0C0} 
\textbf{SSM-Accelerator} & \textbf{H3-SSM} & \textbf{Mamba-SSM} & \textbf{Liquid-S4} \\ \hline
VGA\cite{VGA} & 4\(\times\) & - & - \\ \hline
MARCA\cite{li2024marca} & - & \textbf{11.66}\(\times\) & - \\ \hline
FastMamba\cite{FastMamba} & - & 6.06\(\times\) & - \\ \hline
\acceleratorname & \textbf{3860\(\times\)} & 4.75\(\times\) & \textbf{2000\(\times\)} \\ \hline
\end{tabular}
\end{center}
\end{table}

\subsubsection{PE Utilization}\label{UTIL}
\begin{figure}[th]
    \centering
    \includegraphics[width=0.49\textwidth]{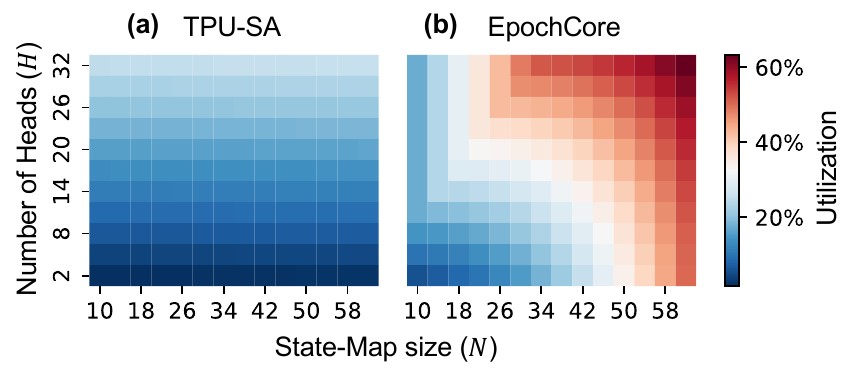}
    \vspace{-0.3in}
    \caption{PE utilization when evaluating an S4 layer using (a) 2-D \(64\times64\) SA, TPU (b) \(64\times64\) \acceleratorname with \newPE for different number of heads \(H\), and state-map size \(N\).}
    \label{fig:SSM_UTIL}
    \vspace{-0.25in}
\end{figure}

\revD{The PE utilization is defined as the ratio of PEs actively engaged in computation to the total number of PEs, under the assumption that no scalability mechanisms—such as scale-up or scale-out—are employed.}
Given a fixed 2-D SA size of \(64\times64\), the PE utilization was calculated for various state-map sizes and number of heads, on TPU-SA and \acceleratorname as shown in Figure~\ref{fig:SSM_UTIL}. 
For TPU-SA, Layer-I and Layer-II of S4 are evaluated in separate SA cycles. 
The PE utilization is taken as the average of the two. 
For \acceleratorname, both Layer-I and Layer-II are performed in the same SA cycle, thus showing improved PE utilization. 
The higher utilization of \acceleratorname over TPU-SA leads to overall better performance, efficiency and scalability of the \acceleratorname.
The PE utilization for Boriakoff's 1-D SA is a constant \(66.7\%\) ~\cite{FFT_SA}. 

%% file: related_work.tex
\section{Related Work}\label{RELATED_WORK}

\begin{table}[]
\caption{Comparison of SSM accelerators}
\label{tab:ssm-accel}
\begin{tabular}{lllll}
\hline
\rowcolor[HTML]{C0C0C0} 
\textbf{Metric} & \textbf{FastMamba} & \textbf{VGA} & \textbf{Marca} & \textbf{EpochCore} \\ \hline
SSM & Mamba2 & H3 & Mamba & S4/Liquid-S4 \\ \hline
Time-Variant & Yes & No & Yes & Yes \\ \hline
Complex value & No & Yes & No & Yes \\ \hline
Speed up over GPU & 8.9X & 14.9X & 11.6X & 2000X \\ \hline
\end{tabular}
\end{table}
Recent advances in structured SSMs have led to a surge of hardware accelerators (MARCA~\cite{li2024marca}, VGA~\cite{VGA}, FastMamba~\cite{FastMamba}) tailored to support long-context sequence modeling. These accelerators vary in their choice of computational paradigm—recurrence vs. convolution—and in their specialization toward specific SSM variants. We categorize and compare these efforts in relation to our \acceleratorname.

\revC{FastMamba~\cite{FastMamba} is a recurrence-based FPGA accelerator tailored for real-valued Mamba models, using quantization-aware co-design techniques—such as Hadamard filtering, power-of-two quantization, and linear activation approximations—to achieve high efficiency via vector processing and fixed-point arithmetic. However, it lacks support for complex-valued SSMs like S4 and H3, limiting its generality for SSM workloads.}

\revC{VGA~\cite{VGA} is a convolution-based accelerator optimized for H3-style global models and long-sequence batch inference. It introduces on-the-fly Vandermonde matrix generation to reduce memory bandwidth and SRAM usage. While it achieves strong speedups and area efficiency over GPUs—especially in memory-bound H3 inference—it remains fundamentally constrained by the offline nature of convolution evaluation.}

MARCA~\cite{li2024marca} optimizes for Mamba-style SSMs by blending linear and element-wise accelerations in a shared PE fabric, heavily reusing units for efficiency, and introducing buffer strategies tailored to Mamba's structure. 

\acceleratorname is the first unified accelerator supporting multiple structured SSMs (S4, Liquid-S4, H3, Mamba) and GEMM-based DNN layers. Unlike convolution-based designs, \acceleratorname exploits the recurrent structure of SSMs via novel \newPE modes and a programmable dataflow (\dataflowname). It supports real and complex MACs, exponential decay primitives, and weight-stationary data reuse, enabling 2000\(\times\) speedup over GPUs for S4 and Liquid-S4 models.

Compared to VGA and MARCA, \acceleratorname is more extensible: gated Mamba variants can be supported through incremental \newPE and \dataflowname enhancements. Its generality and high efficiency make it well-suited for diverse SSM and hybrid deep model workloads.

%% file: conclusion.tex
\vspace{-0.1in}
\section{Conclusion}

This paper presents \acceleratorname, a novel digital accelerator for S4 and Liquid-S4 inference, as well as general DNN workloads. We introduce two key innovations: the \newPE, a specialized processing element supporting real and complex recurrent operations, and \dataflowname, a programmable dataflow optimized for SSMs and GEMM layers.

\acceleratorname achieves up to 250\(\times\) speedup and 45\(\times\) energy savings over Sparse SA accelerators for Liquid-S4 models, and 25\(\times\) faster and 10\(\times\) more energy-efficient than Boriakoff’s 1D SA. Compared to GPUs, \acceleratorname delivers ~2000\(\times\) performance gains on LRA datasets using S4 and Liquid-S4 models, while reducing memory bandwidth usage by 3\(\times\). On other structured SSM models such as H3 and Mamba, \acceleratorname shows performance gains of 3860\(\times\) and 4.75\(\times\) respectively over GPUs.

\newPE incurs a 1.4–2\(\times\) area and up to 1.6\(\times\) power overhead versus traditional PEs, but gated-clock design enables 1.6–3\(\times\) power savings across operating modes. Latency analysis further shows that \acceleratorname maintains balanced execution between S4/Liquid-S4 and DNN layers, scaling efficiently with sequence length—unlike GPU-based systems where S4/Liquid-S4 dominates runtime.